\documentclass[10pt,journal,compsoc]{IEEEtran}
\usepackage{caption}% http://ctan.org/pkg/caption
\usepackage{algorithm}
\usepackage{mathtools}
\usepackage{amsmath,amsfonts,amssymb}
\usepackage{algorithm}
\usepackage{algpseudocode}
\usepackage{filecontents}
\usepackage{multirow}
\usepackage[utf8]{inputenc}
\usepackage[english]{babel}
\usepackage{multirow}

\usepackage[font=footnotesize]{caption}

\algnewcommand\algorithmicinput{\textbf{Input:}}
\algnewcommand\Input{\item[\algorithmicinput]}
\algnewcommand\algorithmicoutput{\textbf{Output:}}
\algnewcommand\Output{\item[\algorithmicoutput]}

\newcommand\numberthis{\addtocounter{equation}{1}\tag{\theequation}}

\usepackage{amsmath}
\usepackage{graphicx}

\usepackage{bm}

\usepackage{tikz}

\usetikzlibrary{shapes.geometric}

\tikzstyle{state} = [ellipse, minimum width=1cm, minimum height=1cm, text centered, draw=black]
\tikzstyle{io} = [ellipse, minimum width=1cm, minimum height=1cm, text centered, draw=white]
\tikzstyle{box} = [rectangle, minimum width=2cm, minimum height=1cm, text centered, draw=white]
\tikzstyle{arrow} = [thick,->,>=stealth]

\makeatletter
\newcommand\footnoteref[1]{\protected@xdef\@thefnmark{\ref{#1}}\@footnotemark}
\makeatother

\usepackage{listings}
\usepackage{xcolor}
\definecolor{orange}{rgb}{0.93, 0.17, 0.17}
\definecolor{noise_color}{rgb}{0.0, 0.0, 1}%{0.94,0.66,0.02}
\definecolor{state1_color}{rgb}{0.0, 0.38, 0.11}%{0.93,0.42,0.65}
\definecolor{state2_color}{rgb}{0.58, 0.0, 0.83}%{0,0.55,0.55}

\usepackage{url}
\newcommand{\specialcell}[2][c]{%
  \begin{tabular}[#1]{@{}c@{}}#2\end{tabular}}

\ifCLASSOPTIONcompsoc
  \usepackage[nocompress]{cite}
\else
  \usepackage{cite}
\fi
\ifCLASSINFOpdf
\else
\fi
\begin{document}
\title{Summarizing Event Sequences with Serial Episodes: A Statistical Model and an Application }
%\title{`` What are the Good Words ? ''-A Frequent Episodes Approach to Dictionary Learning for Document Classification}

\author{Soumyajit~Mitra, %~\IEEEmembership{Member,~IEEE,}
        and~P~S~Sastry %~\IEEEmembership{Fellow,~OSA,}
        %and~Jane~Doe,~\IEEEmembership{Life~Fellow,~IEEE}% <-this % stops a space
\IEEEcompsocitemizethanks{\IEEEcompsocthanksitem Soumyajit Mitra was at the Department of Electrical Engineering, Indian Institute of Science, Bangalore, India. He is currently with Samsung R\&D, Bangalore, India.\protect\\ E-mail: soumyajitmitra93@gmail.com 
\IEEEcompsocthanksitem P.S. Sastry is with the Department
of Electrical Engineering, Indian Institute of Science, Bangalore, India.\protect\\
E-mail: sastry@iisc.ac.in}% <-this % stops an unwanted space
}

\IEEEtitleabstractindextext{%
\begin{abstract}
In this paper we address the problem of discovering a small set of frequent serial episodes from sequential data so as to adequately characterize or summarize the data. We discuss an algorithm based on the Minimum Description Length (MDL) principle and the algorithm is a slight modification of an earlier method, called CSC-2. We present a novel generative model for sequence data containing prominent pairs of serial episodes and, using this, provide some statistical justification for the algorithm. We believe this is the first instance of such a statistical justification for an MDL based algorithm for summarizing event sequence data. We then present a novel application of this data mining algorithm in text classification. By considering text documents as temporal sequences of words, the data mining algorithm can find a set of characteristic episodes for all the training data as a whole. The words that are part of these characteristic episodes could then be considered the only relevant words for the dictionary thus resulting in a considerably reduced feature vector dimension. We show, through simulation experiments using benchmark data sets, that the discovered frequent episodes can be used to achieve more than four-fold reduction in dictionary size without losing any classification accuracy.
\end{abstract}

}

\maketitle

\IEEEdisplaynontitleabstractindextext

% \fi
\IEEEpeerreviewmaketitle

\IEEEraisesectionheading{\section{Introduction}\label{sec:introduction}}

%\IEEEPARstart{T}{his} demo file is intended to serve as a ``starter file''
%for IEEE Computer Society journal papers produced under \LaTeX\ using
%IEEEtran.cls version 1.8b and later.
%% You must have at least 2 lines in the paragraph with the drop letter
%% (should never be an issue)
%I wish you the best of success.

\IEEEPARstart{F}{requent} pattern mining is an important problem in data mining with applications in diverse domains \cite{charu_agarwal_book}. Frequently occurring local patterns can capture useful aspects of the semantics of the data. However, in practice, 
the mined frequent patterns are often large in number and  quite redundant in nature which makes it difficult to effectively use them. 
%In addition, the mining algorithms need a user-set frequency threshold. 
Isolating a small set of non-redundant informative frequent patterns that best describes the data, is an interesting current research problem~\cite{krimp,long_short,tattikdd2011,lam_2014,ibrahim_episodes,vreeken_2017,tkde_news_themes}.
In this paper we are concerned with mining of sequential data in the framework of frequent episodes~\cite{mannila_1997}. We address the problem of isolating a small set of  non-redundant serial episodes that best characterize the data.  
%We also present a statistical generative model to justify our algorithm. 
%In addition, we present an interesting application of this method in text classification. 
	
%Most frequent pattern mining algorithms require an user-defined frequency threshold. In addition, the algorithms suffer from the drawback that the mined patterns are large in number and also quite redundant in nature, which makes it difficult to use the mined patterns to gain useful insights into the data or to use them to extract rules for prediction,classification etc. The primary objective for any pattern mining algorithm is to find a small set of local non-redundant informative frequent patterns that best describes the data.
	
	There have been many recent efforts for extracting a small subset of non-redundant characteristic patterns. There are mainly two families of methods. One family of methods retain only those patterns which are, in some sense, statistically significant. The statistical significance is assessed using either a suitable null model in a hypothesis testing framework or by fitting a generative model for the data source~\cite{gwadera_1,tatti_2009,gwadera_2,lowkam,laxman_hmm,tattikdd2011}. While this can reduce the number of frequent patterns to some extent, this approach cannot tackle redundancy in the discovered patterns.  

Another prominent family of methods for deciding which subset of patterns best explains the data, is based on an information theoretic approach called \textit{Minimum Description Length}~(MDL) principle~\cite{mdl_book}. In the context of the problem of isolating  a `best' subset of frequent patterns, the use of MDL principle can be explained as follows. We formulate a mechanism so that given any subset of frequent patterns we can use them as a `model' to encode the data. Then, the subset that results in the overall best level of data compression is considered to be the subset that best characterizes the data. Such a view, motivated by MDL principle, has been found effective for many frequent pattern mining algorithms~\cite{mdl_ch_charu_book}. 

MDL principle views learning as data compression. If we are able to discover all the important regularities in data then we should be able to use these to compress the data well. In this view, the coding mechanism used should be lossless; that is the original data should be exactly recoverable given the encoded compressed representation.

% In this paper we adopt an existing algorithm based on such use of the MDL principle. Our main contribution is in developing  a novel generative model using which we can provide statistical justification for the algorithm.  
	
	The Krimp algorithm~\cite{krimp} is one of the first methods that used MDL principle to identify a small set of relevant patterns in the context of frequent itemset mining. As mentioned earlier, in this paper we are concerned with sequential data. For sequential data, unlike in the case of transaction data, the temporal ordering of data tuples is important and our encoding mechanism should be such that we should be able to recover the original data sequence in correct order along with all time stamps. This presents additional complications while encoding sequential data using frequent patterns. 
(See~\cite{ibrahim_episodes} for more discussion on this).  
%It is seen to be quite effective in isolating a small set of useful frequent itemsets. 
There are many MDL-motivated algorithms proposed for characterizing sequence data through a subset of frequent patterns~\cite{long_short,lam_2014,ibrahim_episodes,vreeken_2017}. Different algorithms use different strategies for coding data using frequent patterns. While the methods are motivated by MDL principle, the coding strategies and hence the computation of compression achieved by a given subset of frequent patterns are essentially arbitrary. %While some of these algorithms achieve higher level of data compression than others, it is not clear whether that is in any way significant. This is because the algorithms do not have any clear statistical basis. 
%Under the MDL principle, we are interested in lossless compression of the data. That is, when we use a set of patterns as a model to encode the data, the coding should be such that we should be able to recreate the complete data without any errors. 
%For sequential data, unlike in the case of transaction data, the temporal ordering of data tuples is important and this presents additional complications while encoding the data using frequent patterns. 
%(See~\cite{ibrahim_episodes} for more discussion on this).  
%For example, in the context of frequent episode mining in event sequences, the events constituting an episode occurrence need not be contiguous and hence one needs to have some way of taking care of the inter-event gaps while encoding the data. In \cite{long_short,lam_2012,lam_2014}, they explicitly record such gaps while encoding the data, thus significantly increasing the encoding length. In some cases, the resulting encoding may become even longer than the raw data \cite{lam_2012} which is, philosophically, not satisfying under the MDL principle. 
%This is not useful, as the underlying philosophy of MDL suggests that one needs a good level of data compression to have confidence in the model. T

In this paper we consider a recently proposed algorithm called CSC-2~\cite{ibrahim_episodes} which is an efficient method to discover a subset of serial episodes that best characterizes data of event sequences. It uses a novel pattern class consisting of injective serial episodes with fixed inter-event times. A similar pattern class was also used recently for learning association rules from temporal data~\cite{xiang_2018}. 
%In this paper we attempt to provide a statistical justification 
%Recently, Ibrahim et al~\cite{ibrahim_episodes} presented an efficient method, called CSC-2, for isolating a subset of characteristic patterns. They use a novel pattern class consisting of injective serial episodes with fixed inter-event times. A similar pattern class was also used recently for learning association rules from temporal data~\cite{xiang_2018}.
%This method  does not need to explicitly encode the inter-event gaps in an episode occurrence while still giving lossless compression. 
%This paper also addresses this problem of discovering a small set of frequent episodes that best characterizes sequential data. 
%In this paper, we use this method. 
The CSC-2 algorithm uses the number of distinct occurrences of an episode as its frequency. Here, we extend it to the case of non-overlapped occurrences as episode frequency and then provide some statistical justification for the algorithm based on a generative model. 
%An attractive feature of the algorithm is that it does not need any user-specified thresholds. 

The main contribution of this paper is a HMM-based generative model which provides some statistical justification for the CSC-2 algorithm. 
In all MDL-based approaches, a subset of patterns is selected based on the data compression it can achieve. This depends on the (arbitrary) coding scheme used by the algorithm which is selected heuristically. In this paper we provide a justification for the coding scheme and the algorithm used in CSC-2 based on our proposed statistical generative model. This is the first time, to our knowledge, that such a formal connection is established between mining of episodes using the MDL principle and a generative model for data source.  Since this generative model is Markovian and hence can handle only non-overlapped occurrences based episode frequency, we extended CSC-2 to use non-overlapped occurrences as episode frequency.

% One of the earliest generative models for frequent episodes is the so called Episode Generating HMMs introduced in~\cite{laxman_hmm}. This is a generative model for data consisting of many non-overlapped occurrences of a single episode and was used essentially to assess the statistical significance of any one given serial episode. Motivated by this, we present a HMM based generative model for pairs of episodes. For such a generative model to be reasonable, it should generate data consisting of many (non-overlapped) occurrences of the two episodes. While two occurrences of the same episode should be  non-overlapped, it should also be possible for occurrences of different episodes to have arbitrary overlap. Accommodating this makes our HMM model much more complex than the one  in~\cite{laxman_hmm}. While the model we present is only for a pair of episodes, we feel that the conclusions drawn from it are adequate for the purpose of justifying the main idea in our MDL based algorithm for discovering a subset of characterizing episodes. The reason we use non-overlapped occurrences as the frequency of episodes in our algorithm is that for non-overlapped occurrences one can build such Markovian generative models.

Another major contribution of this paper is a novel application of this method of discovering a set of characteristic episodes from sequential data. The application is in text classification. Most text classification methods represent each document as a vector over a dictionary of words which is often called the bag-of-words representation. The dictionary for this is taken to be all the words in the corpus (after appropriate stemming and dropping of stop words). Often, the dictionary sizes are large resulting in high dimensionality of the feature vectors representing individual documents. A text document can be viewed as a sequence of events with event types being words. Hence, using our method, we can discover a subset of characteristic episodes that best represent the full corpus of document data. We can then use the words (event-types) in the subset of discovered episodes to form our dictionary. Since our method does not even need a frequency threshold, this constitutes a parameter-less unsupervised method of feature selection for this problem. We show, through empirical experiments, that this method results in a very significant reduction of dictionary size without any loss of classification accuracy. 

The rest of the paper is organized as follows. Section II describes the episode mining algorithm. Section III presents our proposed generative model. Section IV explains our method of finding a smaller sized dictionary in text classification problems and reports results obtained on different text datasets. Conclusions are presented in Section-V. 

\section{Discovering Best Subset of Serial Episodes}\label{sec:algorithm}
\subsection{Episodes in Event Sequences}
We begin with a brief informal description of the episodes framework. (See \cite{mannila_1997,achar_survey} for more details). Here the data is (abstractly) viewed as an event sequence denoted as $\mathcal{D}= <(E_1,t_1),(E_2,t_2),...,(E_n,t_n)> $ where, in each tuple or {\em event}, $(E_i,t_i)$, $E_i$ is the event-type and $t_i$ is the time of occurrence of that event. We have $E_i \in \mathcal{E}$, a finite alphabet set and $t_i \leq t_{i+1}$, $\forall i$. An example event sequence is
	\begin{equation}
		\label{eq:1}
		\begin{split}
			\mathcal{D}\; = \ <(D,1),(A,2),(C,3),(E,3),(A,4),(B,4),(C,5), \\
			(D,5),(B,6),(C,7),(E,7),(A,7),(C,8),(B,8),(C,9)>
		\end{split}
	\end{equation}
The patterns of interest here are called episodes. In this paper we are concerned with only serial episodes. 
%\noindent
%	\textbf{Definition 1.} \textit{[Mannila et al., 1997] An N-node serial episode $\alpha$, is a tuple, $(V_{\alpha}, <_{\alpha}, g_{\alpha})$, where $V_{\alpha} = \{ v_{1},v_{2},... ,v_{N} \}$, $<_{\alpha}$ is a total order on $V_{\alpha}$ and $g_{\alpha}: V_{\alpha}\to \mathcal{E}$.} \newline
%A serial episode is an ordered sequence of event types. 
%Without loss of generality we assume $v_1 <_{\alpha} v_2 <_{\alpha} \cdots$ and denote $g_{\alpha}(v_i) = \alpha[i]$. Hence, 
We represent an $N$-node serial episode, $\alpha$, as $\alpha[1] \rightarrow \cdots \rightarrow \alpha[N]$ where $\alpha[i]$ is the event-type of the $i^{th}$ event of the episode. An episode is said to be injective if all event types in the episode are distinct. For example, $A \rightarrow B \rightarrow C$ is a three node injective serial episode.
%\footnote{In this paper we would be concerned only with injective episodes. Hence, unless otherwise specified, in the rest of the paper when we say serial episode we mean an injective serial episode.} 
An occurrence of the serial episode is constituted by events in the data sequence that have appropriate event types and their times of occurrence are in the correct order. In (\ref{eq:1}), $( (A,2), (B,4), (C,5) )$ constitutes an occurrence of $A \rightarrow B \rightarrow C$ while $( (A,4), (B,4), (C,5) )$ does not because $B$ does not occur {\em after} $A$. (Note that the events constituting an occurrence need not be contiguous in the data). 

The data mining problem is to discover all frequently occurring episodes. In the frequent episodes framework, many different frequency measures are defined based on counting different subsets of occurrences. We mention two such frequencies below which are relevant for this paper. There are efficient algorithms for discovering serial episodes under many frequency measures. (See \cite{achar_survey} for more details on different frequencies and algorithms for discovering serial episodes). 

Two occurrences of a serial episode are said to be non-overlapped if no event of one occurrence is in between events of the other. In (\ref{eq:1}), $( (A,2), (B,4), (C,5) )$ and $( (A,7), (B,8), (C,9) )$ are non-overlapped occurrences of $A \rightarrow B \rightarrow C$ while $( (A,4), (B,6), (C,8) )$ is another occurrence of this episode which overlaps with both the earlier ones.  The \textbf{non-overlapped frequency} of an episode is defined as the maximum number of non-overlapped occurrences of the episode in the event sequence~\cite{laxman_hmm}. Two occurrences are said to be distinct if they do not share any event. All three occurrences above are distinct.
The maximum number of distinct occurrences is another frequency of interest. 
 %(See \cite{achar_survey} for more details on different frequencies and algorithms for discovering serial episodes).  

In our method here we use a special class of serial episodes called {\em fixed-interval serial episodes}~\cite{ibrahim_episodes}. A fixed interval serial episode can be denoted as $\alpha= (\alpha[1] \xrightarrow{\Delta_{1}} \alpha[2] \xrightarrow{\Delta_{2}} \cdot \cdot \cdot \xrightarrow{\Delta_{N-1}} \alpha[N])$ where $\Delta_i$ is the prescribed gap between the times of $i^{th}$ and $(i+1)^{st}$ events of any occurrence of $\alpha$. For example, $A\xrightarrow{2}B\xrightarrow{1}C$ is a fixed interval injective serial episode. In (\ref{eq:1}), $( (A,2), (B,4), (C,5) )$ is an occurrence of this episode while $( (A,7), (B,8), (C,9) )$ is not.
 
As is easy to see, all events constituting an occurrence of a fixed interval serial episode are completely specified by giving only the time of occurrence of the first event. 
Also, two occurrences starting at different times would be distinct if the episode is injective (that is, all event types in the episode are different).  

%	\textbf{Definition 3.} \textit{An \textbf{occurrence} of an episode $\alpha=(V_{\alpha}, <_{\alpha}, g_{\alpha})$, in a sequence $\mathcal{D}=<(E_1,t_1),(E_2,t_2),...(E_n,t_n)>$, is an injective mapping $h : V_{\alpha}\longrightarrow \{ 1,...,n \}$ such that $g_{\alpha}(v)=E_{h(v)}, \forall v \in V_{\alpha}$, and for all $u,v \in V_{\alpha}$ with $u\neq v$ and $u <_{\alpha} v$ we have $t_{h(u)}<t_{h(v)}.$} \newline
%	\textbf{Definition 4.} \textit{[Laxman et al., 2005] Two occurrences of an episode are said to be non-overlapping if no event associated with one appears in between the events associated with other. The \textbf{non-overlapping frequency} of an episode is defined as the maximum number of non-overlapping occurrences of the episode in the event sequence.} \newline
%	\textbf{Definition 5.} \textit{Two occurrences, $h_1, h_2$ of an episode are said to be \textbf{distinct} if none of the events of one occurrence is among events of the other. i.e., $h_1(u)\neq h_2(v)$, $\forall u,v \in V_{\alpha}$. The distinct occurrence frequency of an episode $\alpha$ is the cardinality of a maximal set of distinct occurrences of $\alpha$.}\newline
%	\textbf{Definition 6.} \textit{The (lexicographic) ordering, $<_\star$, on $\mathcal{H}$, the set of all occurrences of an episode is defined as: for any two occurrences $h_1$ and $h_2$ of that episode, we say $h_1 <_\star h_2$ if the least i for which $t_{h_1(v_i)}\neq t_{h_2(v_i)}$ is such that $t_{h_1(v_i)} <  t_{h_2(v_i)}$. This is a total order on the set $\mathcal{H}$.} 
	\subsection{Mining Algorithm for the Best Subset}
Here our interest is in discovering a small set of fixed-interval serial episodes that best explains the data. We use the Minimum Description Length~(MDL) principle for this. Hence we rank different subsets of episodes by the total encoding length that results when we use them as models to encode data. Under MDL, the encoding should be such that we should be able to recover the original data completely. Since we are considering sequential data, this means we should be able to recover the data in the original sequence with all time stamps. 

We first explain the strategy of coding the data sequence using our episodes. The basic idea is that we can encode all events constituting the occurrence of a fixed interval serial episode by just giving the start times of the occurrence. The encoding strategy is same as that used in~\cite{ibrahim_episodes}. For obtaining the best subset of episodes we essentially use the CSC-2 algorithm from~\cite{ibrahim_episodes} with the main difference being we use the non-overlapped frequency while that algorithm uses distinct occurrences as frequency. Below we first explain the encoding scheme through an example and then briefly explain the CSC-2 algorithm. (For more details on the encoding scheme and the CSC-2 algorithm, please see~\cite{ibrahim_episodes}). 

Table \ref{table:1} illustrates the coding scheme by encoding the event sequence in \eqref{eq:1} using essentially three arbitrarily selected episodes. Each row specifies the size and description of an episode, the number of occurrences of the episode and the start times of these occurrences. Thus, the first row of Table~\ref{table:1} specifies a three-node episode, namely, $(A\xrightarrow{2}B\xrightarrow{1}C)$, which has two occurrences starting at time instants 2 and 4. Thus, this row codes for six events in the data constituting the two occurrences of this 3-node episode. Similarly the second row codes for six events by specifying two occurrences of a 3-node episode and the third row codes for two events by specifying one occurrence of a 2-node episode. Suppose we are interested in asking how good is this subset of three episodes. These three episodes together, as specified through Table~\ref{table:1}, account for all but two events in the data. But coding under MDL should be lossless. Hence, in the last row of Table~\ref{table:1} we have used two occurrences of a 1-node episode to make sure that all events in the data sequence are covered.  It is easy to see that given this table, we can recreate the entire data sequence exactly. In this table we can think of the first two columns as coding the model, that is the subset of episodes, and the last two columns as coding the data using this model. Thus the length or size of this table can be the total encoded length for the subset of episodes. Given any subset of episodes (such as the three episodes in the first three rows of the table) we can find an encoding like this for the whole data by adding occurrences of a few 1-node episodes as needed (which is what is done in the fourth row of the table). 

 In this table, one can see that the event $(C,5)$ is coded for, by both the first and the second  episode in the table. Intuitively, we get better data compression if such overlaps among the parts of data encoded by different episodes in the selected set, are minimized. Thus, we should get better compression of data if we can choose episodes with high frequency (so that they can cover for large number of events) which are non-redundant (so that the overlaps as mentioned above are reduced). This is the intuitive reason for using this coding scheme and looking for a subset of episodes that achieves best compression of data.  
	\begin{table}[h]
		\centering
		\caption{Encoding of event sequence}
		\label{table:1}
		\resizebox{8cm}{!}{
			\begin{tabular}{|c|c|c|c| } 
				\hline
				\textbf{Size of episode} & \textbf{Episode name}  & \textbf{No. of occurrences} & \textbf{List of Occurrences} \\ \hline
				$3$ & $(A\xrightarrow{2}B\xrightarrow{1}C)$ & $2$ & $<2,4>$ \\ 
				$3$ & $(D\xrightarrow{2}E\xrightarrow{2}C)$ & $2$ & $<1,5>$ \\ 
                                $2$ & $(A\xrightarrow{1}B)$ & $1$ & $<7>$ \\
				$1$ & $C$ & $2$ & $<3,8>$ \\ \hline
		\end{tabular}}
	
	\end{table}
%	As is easy to see, each episode description (namely its event types and its inter-event gaps) along with the start times of every occurrence enables exact decoding of all the events constituting the occurrences of that episode. The information provided in Table \ref{table:1} is sufficient to decode the data sequence in \eqref{eq:1}. 
	 
Our objective is to find a subset of episodes to encode data like this so as to get best data compression. For purposes of counting length/memory we assume that event types as well as times of occurrence are integers and that each integer takes one unit of memory. Let $\alpha$ be an N-node episode of frequency $f_{\alpha}$ used for encoding. Its row in the table would need $2N+1+f_{\alpha}$ units ($1$ unit to represent the size of the episode, $N$ units to represent the event-types of the episode, $N-1$ units for representing the inter-event gaps, $1$ unit for frequency and $f_{\alpha}$ units to represent the start times of the occurrences). Since non-overlapped or distinct occurrences do not share events, this episode encodes for $f_{\alpha}N$ events in the data and hence we need at least $f_{\alpha}N$ units of memory if we want to encode these events in the data using 1-node episodes. Define

%	To state formally, let $\alpha$ be a N-node episode of frequency $f_{\alpha}$. In case of distinct or non-overlapping frequency, no event is shared between two occurrences of the episode. Hence, the occurrences of episode $\alpha$ covers $f_{\alpha}N$ number of events in the event sequence and hence minimum number of units required to encode these events using only singletons (trivial encoding) is $f_{\alpha}N$. On the other hand, using the above encoding scheme, number of units required is $2N+1+f_{\alpha}$ ($1$ unit to represent the size of the episode, $N$ units to represent the event-types of the episode, $N-1$ units for representing the inter-event gaps, $1$ unit to encode the frequency of the episode and $f_{\alpha}$ units to represent the start time of the occurrences).

%	Keeping in mind the encoding scheme, we state the following metric.
	\begin{equation}
		\label{eq:2}
		score(\alpha, \mathcal{D}) =\big(  f_{\alpha}N \big) - \big(2N +1 + f_{\alpha}  \big)
	\end{equation}
	
	If $score(\alpha,\mathcal{D})>0$, then we can conclude that $\alpha$ is a \textit{useful candidate}, since, selecting it can improve encoding length (in comparison to the trivial encoding using only 1-node episodes). However, the true utility of $\alpha$ is to be assessed with respect to what it would add to compression given the other selected episodes. 
	
%	But $score$ alone is not sufficient to declare an episode interesting. We want the selected episodes to be non-redundant as well. 
Let $\mathcal{F}_{s}$ be a set of episodes of size greater than one. Given any such $\mathcal{F}_{s}$, let $L(\mathcal{F}_{s},\mathcal{D})$ denote the total encoded length of $\mathcal{D}$, when we encode all the events which are part of the occurrences of episodes in $\mathcal{F}_{s}$, by using episodes in $\mathcal{F}_{s}$ and encode the remaining events in $\mathcal{D}$, if any, by episodes of size one. Given any two episodes $\alpha\text{, }\beta$, let $OM(\alpha,\beta)$ denote the number of events in the data that are covered by occurrences of both $\alpha$ and $\beta$ in the data sequence $\mathcal{D}$. Define
	\begin{equation}
	%	\label{eq:3}
		\resizebox{8cm}{!}{
			$ overlap\text{-}score(\alpha,\mathcal{D},\mathcal{F}_{s}) = score(\alpha,\mathcal{D}) - \textstyle\sum\limits_{\beta \in \mathcal{F}_{s}}OM(\alpha,\beta)$}
\label{eq:os}
	\end{equation}
	$Overlap\text{-}score$ gives an estimate of how much extra encoding efficiency can be achieved by selecting $\alpha$ given the set $\mathcal{F}_{s}$.
	It can be proved~\cite[Prop.~1]{ibrahim_episodes} that 
	\begin{equation*}
		\resizebox{9cm}{!}{
			$ \text{If } overlap\text{-}score(\alpha,\mathcal{D},\mathcal{F}_{s})>0, \text{then } L(\mathcal{F}_{s} \cup \{ \alpha \},\mathcal{D})<L(\mathcal{F}_{s},\mathcal{D})$}
	\end{equation*}
	This means that, given a current set of episodes $\mathcal{F}_{s}$, adding to $\mathcal{F}_{s}$ an episode $\alpha$ with positive $overlap\text{-}score$, would only reduce the total encoded length. 
The CSC-2 algorithm in~\cite{ibrahim_episodes} is essentially a greedy algorithm that keeps adding episodes with highest $overlap\text{-}score$. 
%Thus, the greedy heuristic of CSC-2 algorithm is to select the episode with maximum overlap-score. (Note that by definition, $overlap\text{-}score(\alpha,\mathcal{D}, \mathcal{F}_s ) = score(\alpha,\mathcal{D}) \text{ if }\mathcal{F}_s = \phi)$. 
This greedy selection of best episode (based on $overlap\text{-}score$) is done from a set of candidate episodes, generated through a depth-first search of the lattice of all serial episodes. Each candidate episode is the `best' episode in one of the paths of the depth-first search tree. For the sake of completeness we give the pseudocode of this algorithm as Algorithm~\ref{algo:1} (For more details see \cite{ibrahim_episodes}).

	\begin{algorithm} [h]
		\caption{CSC-2($\mathcal{D},T_g,K$)}
		\label{algo:1}
		\begin{algorithmic}[1]
			\Input{
				$\mathcal{D}$: Event sequence, $T_g$: maximum inter-event gap, $K$: maximum number of selected episodes.
			}
			\Output{The set of selected frequent episodes $\mathcal{F}$.}
			\State Initialize the final set of selected episodes $\mathcal{F}$ as $\emptyset$
		%	\State $coveringexists \gets true$
			\While {\textit{data compression achievable} and $|\mathcal{F}| < K$} 
			\State $\mathcal{F}_s \gets \emptyset$
			\State $\mathcal{C} \gets$ Generate Candidate Episodes %\textsc{BestExtensions}$(\mathcal{D},T_g)$
			\State Calculate events shared by occurrences for every pair of candidate episodes %\textsc{FindOverlapMatrix}$(\mathcal{D},\mathcal{C})$
			\Repeat
			\State $\alpha \gets \mathrm{argmax}_{\gamma \in \mathcal{C}} overlap$-$score(\gamma, \mathcal{D}, \mathcal{F}_s)$
			\If {$overlap$-$score(\alpha, \mathcal{D}, \mathcal{F}_s)>0$} \
			\State$\mathcal{F}_s \gets \mathcal{F}_s \cup \{\alpha\}$
			\State $\mathcal{C} \gets \mathcal{C} \backslash \alpha$
			%\ElsIf {$overlap$-$score(\alpha, \mathcal{D}, \mathcal{F}_s) \leq 0$ \textbf{and} $\mathcal{F}_s = \emptyset$ } 
			%\State $coveringexists \gets false$
			\EndIf
			\Until {$overlap$-$score(\alpha,\mathcal{D},\mathcal{F}_s) \leq 0$ \textbf{or} $|\mathcal{F} \cup \mathcal{F}_s|=K$}
			%\State $\mathcal{D} \gets \mathcal{D} \backslash (occurrences$ $of$ $\mathcal{F}_s)$
			\State Delete the events from $\mathcal{D}$ corresponding to the occurrences of selected episodes $\mathcal{F}_s$
			\State $\mathcal{F} \gets \mathcal{F} \cup \mathcal{F}_s$
			\EndWhile
			%\State $\mathcal{A} \gets$ Size-1 episodes in remaining $\mathcal{D}$
			\State $\mathcal{A} \gets$ Size-1 episodes to encode remaining events in $\mathcal{D}$
			\State $\mathcal{F} \gets \mathcal{F} \cup \mathcal{A}$
			\State \Return {$\mathcal{F}$}
		\end{algorithmic}
	\end{algorithm}
	
We can run this algorithm to find `top-$K$' best episodes. If we give a very large value of $K$, the algorithm exits when it cannot find any more episodes (of size greater than 1) which improves coding efficiency. The algorithm needs no frequency threshold given by users. Our  $overlap$-$score$ naturally prefer episodes with higher frequency and we need no threshold because we pick episodes based on what they add to coding efficiency. Thus, the algorithm does not really have any hyperparameters (except for $T_g$, the maximum allowable inter-event gap which is not a critical one). 
	
While calculating $overlap$-$score$, we need to decide what type of occurrences we would count toward frequency. 
	As mentioned earlier, CSC-2 uses distinct occurrences. In this paper we use non-overlapped occurrences for frequency of episodes. The reason for this is that the generative model we present in the next section is for non-overlapped occurrences. Also, in our application to text classification, non-overlapped occurrences is a more natural choice for frequency. 

We obtain the sequence of non-overlapped occurrences from the distinct occurrences returned by CSC-2 using a simple algorithm. We take the first occurrence from the sequence of distinct occurrence as the first one in the sequence of non-overlapped occurrences. Then onwards we take the first distinct occurrence starting after the last non-overlapped occurrence we have as the next one in our sequence of non-overlapped occurrences. The pseudocode for this algorithm is listed as  Algorithm~\ref{algo:2}.
Below, we prove the correctness of this algorithm. That is, we show that the sequence of non-overlapped occurrences we get is a maximal one and hence we get the correct non-overlapped frequency. 
%We now present an algorithm (Algorithm \ref{algo:2}) to find a maximal set of non-overlapped occurrences of an episode directly from its distinct occurrences and then prove its correctness. 
	
	\begin{algorithm} [h]
		\caption{Find-NO-occurrences($\alpha$)}
		\label{algo:2}
		\begin{algorithmic}[1]
			\Input{
				$OccStarttime\text{-}List(\alpha)$: List of start times of distinct occurrences of $\alpha$ in increasing order of start times, $\Delta_i \quad \forall i \in \{ 1,2,\dots,N-1 \}$: Inter-event gaps of episode $\alpha$.
			}
			\Output{NO-occurrences ($\alpha$): List of start times of a maximal set of non-overlapping occurrences of  $\alpha$.}
			\State $\text{NO-occurrences}(\alpha) \gets \emptyset$ 
			\State $itr:$ pointer to first element of $OccStarttime \text{-}List(\alpha)$
			\If {$OccStarttime\text{-}List$ is empty } \
			\State \Return{NO-occurrences$(\alpha)$}
			\EndIf
			\State $t_s \gets itr.starttime$
			\State $\text{NO-occurrences}(\alpha) \gets \text{NO-occurrences}(\alpha) \cup \{ t_s \}$
			\State $itr \gets itr.next$
			\While{$itr \neq NULL\text{(end of list)}$} 
			\State $t^{'}_s \gets itr.starttime$
			\If{$t^{'}_s > \Bigg( t_s + \sum\limits_{i=1}^{N-1} \Delta_i \Bigg)$ } \
			\State $\text{NO-occurrences}(\alpha) \gets \text{NO-occurrences}(\alpha) \cup \{ t^{'}_s \}$
			\State $t_s \gets t^{'}_s$
			\EndIf
			\State $itr \gets itr.next$
			\EndWhile
			\State \Return{$\text{NO-occurrences}(\alpha)$}
		\end{algorithmic}
	\end{algorithm}
	
	Let $\mathcal{H}=\{ h_1,h_2,\dots,h_l \}$ denote the set of non-overlapped occurrences returned by Algorithm \ref{algo:2}. Each occurrence, $h_i$ can be thought of as a tuple of indices in the data sequence which give the position of events in data that constitute this occurrence. For example, in data sequence~\eqref{eq:1}, the occurrence of the episode $A \rightarrow B \rightarrow C$ constituted by the events $<(A,2), (B,4), (C,5)>$ would be represented by the tuple $(2\; 6 \; 7)$.  Hence, as a notation, we use $t_{h_i(k)}$ to denote the time of the $k^{th}$ event of the episode in the occurrence $h_i$. On the set of occurrences, $\mathcal{H}$, there is a natural order: occurrence $h_i$ is {\em earlier} than $h_j$ if the $t_{h_i(1)} < t_{h_j(1)}$.  Because of the way the occurrences  in $\mathcal{H}$ are selected by our algorithm, the following property is easily seen to hold:\newline
	\textbf{Property 1: }$h_1$ is the {\em earliest} distinct occurrence of the episode $\alpha$. For any $i$, $h_i$ is the first distinct occurrence starting after $t_{h_{i-1}(N)}$ and there is no distinct occurrence which starts after $t_{h_l(N)}$.

%We prove the correctness of Algorithm \ref{algo:2} below. \newline

	\textbf{Proposition 1: } $\mathcal{H}$ is a maximal set of non-overlapped occurrences of $\alpha$ \newline
	\textit{Proof: }Note that for fixed interval injective serial episodes, occurrences having different start times are distinct. Consider any other set of non-overlapped occurrences of the episode, $\mathcal{H}^{'}=\{ h'_1,h^{'}_2,\dots, h^{'}_m \}$. Let $p=min\{ m,l \}$. We first show that 
	\begin{equation}
		\label{eq:4}
		t_{h_i(N)} \leq t_{h^{'}_i(N)} \quad \forall i \in \{ 1,2,\dots,p \} 
	\end{equation}
	We use induction on $i$ to prove this. Let us show this first for $i=1$. Suppose, $  t_{h_1(N)} >t_{h^{'}_1(N)} $. Since, the inter-event gaps are fixed, we have $t_{h_1(1)} >t_{h^{'}_1(1)}$. This means we have found a distinct occurrence of the episode which starts before $h_1$. This contradicts the first statement of Property 1 that $h_1$ is the {\em earliest} distinct occurrence. Hence, $t_{h_1(N)} \leq t_{h^{'}_1(N)}$.
	
	Suppose, $t_{h_i(N)} \leq t_{h^{'}_i(N)}$ is true for some $i<p$. We show that $t_{h_{i+1}(N)} \leq t_{h^{'}_{i+1}(N)}$. Suppose, $ t_{h_{i+1}(N)} > t_{h^{'}_{i+1}(N)} $. This implies $  t_{h_{i+1}(1)} > t_{h^{'}_{i+1}(1)} $. Again, since, $\mathcal{H}^{'}$ is a set of non-overlapped occurrences, we have $ t_{h^{'}_{i}(N)} < t_{h^{'}_{i+1}(1)} $. Hence, we have 
	\begin{equation*}
		t_{h_i(N)} \leq t_{h^{'}_i(N)} <  t_{h^{'}_{i+1}(1)} < t_{h_{i+1}(1)}  
	\end{equation*}
	But this contradicts the fact of Property 1, that $h_{i+1}$ is the earliest distinct occurrence after $t_{h_i(N)}$. Hence, $ t_{h_{i+1}(N)} \leq t_{h^{'}_{i+1}(N)} $.
	
	Now we prove the maximality of the set $\mathcal{H}$. Suppose, we assume that $|\mathcal{H}^{'}| > |\mathcal{H}|, \text{ i.e }, \; m>l$. From inequality \eqref{eq:4}, $h^{'}_{l+1}$ is an occurrence beyond $t_{h_l(N)}$. But this contradicts the last statement of Property 1 that there is no distinct occurrence beyond $t_{h_l(N)}$. Hence, $|\mathcal{H}| \geq |\mathcal{H}^{'}|$ for every set of non-overlapped occurrences $\mathcal{H}^{'}$. This proves the maximality of the set $\mathcal{H}$.
%\blob

We can now sum up our method of finding a subset of serial episodes that best characterizes the data sequence. We use the coding scheme as described here and use a greedy heuristic to find the subset that achieves the best compression. This is essentially the same as the CSC-2 algorithm of~\cite{ibrahim_episodes}.
 However, we use Algorithm~\ref{algo:2} to get non-overlapped occurrences of episodes from distinct occurrences and then use that frequency in selecting episodes with best overlap-score. In the next section we present an interesting generative model that provides some statistical justification for our algorithm based on selecting an episode with best overlap-score. 

\section{A Generative model for Pairs of Episodes}
In this section we present a class of generative models which is a specialized class of HMMs. (This model is motivated by a HMM-based model for single episodes proposed in~\cite{laxman_hmm}). An HMM contains a Markov chain over some state space. But the states are unobservable. In each state, a symbol is emitted from a finite symbol set according to a symbol probability distribution associated with that state. The stream of symbols is the observable output sequence of the model. 

In our case, the symbol set would the set of event-types and thus the observed output sequence would be a sequence of event-types. We think of this as an event sequence where the event-times are not explicitly specified. For occurrences and hence for frequencies of general serial episodes (without any inter-event times specified) only the time-ordering of the event-types in the data sequence is important; actual event times play no role. Hence in this section we consider serial episodes without any fixed inter-event times.

In our generative model, the state transition probability matrix of the Markov chain is parameterized by a single parameter, which is called the noise parameter. For every pair of serial episodes, we have one such generative model.
%\footnote{The model generates a sequence of event types without any explicit time stamps on them. Hence we consider general serial episodes (without any fixed gaps) because occurrence of serial episodes is determined only by the order of event types and not explicitly by the time stamps.} 
For small enough value of the noise parameter, the model is such that the output from the model would be an event sequence containing many non-overlapped occurrences of the two corresponding episodes. While occurrences of any one episode would be non-overlapped in the output event sequence, an occurrence of one episode may be arbitrarily interleaved with occurrences of the other episode. Thus this is a good class of generative model for a data source where a pair of episodes form the most prominent frequent patterns (under the frequency based on non-overlapped occurrences). This is the first instance of such a statistical generative model for multiple episodes. 

%To be able to justify an inference that a subset of episodes best describes some data, one needs a model that properly captures interactions among episodes. The model we present here is a first step in that general direction.  
%As a matter of fact, given the output event sequence of the model, any standard algorithm for frequent serial episode mining~\cite{achar_survey} would find the two episodes to be the significant ones. 

Consider the family of such models containing a model for every possible pair of episodes. Let $\Lambda_{\alpha \beta}$ denote the model for the pair of episodes $\alpha$ and $\beta$. Given an event sequence we can now ask which is the maximum likelihood estimate of a model from this class of models. This would essentially tell us which pair of episodes best `explains' the data sequence in the sense of maximizing the likelihood. We show that such a pair of episodes are not necessarily the two most frequent episodes. The data likelihood depends both on the frequencies of the episodes as well as on the number of events in the data that the occurrences of the two episodes share. Thus, we show, for example, that $\Lambda_{\alpha \beta}$ may have better likelihood than $\Lambda_{\alpha \gamma}$ even when $\beta$ has lower frequency than $\gamma$, if overlap between $\alpha$ and $\beta$ is much less than that between $\alpha$ and $\gamma$.
%pair of episodes which are such that the fraction of events in the event sequence that are part of the occurrences of the two episodes is maximum. 
The results we present here provide some statistical justification for the coding scheme and the algorithm that we presented in the previous section. 

\subsection{The HMM model}
%In this section, we formally connect our episode mining algorithm with learning generative models for the data in the form of a specialized class of HMMs. HMMs are very popular for modeling temporal sequence data with applications to many domains like speech processing and bioinformatics. For the sake of completeness, we give a brief overview of discrete HMMs that we use.

%An HMM contains a Markov chain over some state space. But the states are unobservable. In each state, a symbol is emitted from a finite symbol set according to the symbol probability distribution. The stream of symbols is the observable output sequence of the model. 

A HMM is specified as $\Lambda = (\mathcal{P}, \pi, b)$ where $\mathcal{P} = [p_{ij}]$ is the state transition probability matrix of the Markov chain with state space, say, $S$, $\pi$ is the initial state probabilities and $ b= (b_q, \; q \in S)$ where $b_q$ denotes the symbol probability distribution in state $q$.
Let $\textbf{o} = (o_1,o_2, \cdots, o_T)$ be an observed symbol (or output) sequence. 
%Let $\mathcal{P}$ be the state transition probability matrix, where $p_{ij}$ is the transition probability from state $i$ to state $j$, $\pi$ be the initial state probabilities and $b$ be the symbol probability distributions. 
The joint probability of the output sequence $\textbf{o}$ and a state sequence  $\textbf{q} = (q_1, q_2, \cdots, q_T)$ given an HMM $\Lambda$ is
\begin{equation}
\label{eq:5}
P(\textbf{o,q}| \Lambda) = \pi_{q_1} b_{q_1}(o_1) \prod_{t=2}^{T} p_{q_{t-1}q_{t}}b_{q_t}(o_t)
\end{equation}
To determine the model with maximum likelihood, we need to find $P(\textbf{o}|\Lambda)$. 
This data likelihood is often assessed by evaluating the above joint probability of \eqref{eq:5} along a \textit{most likely} state sequence, $\textbf{q}^{*}$, where 
\begin{equation}
\textbf{q}^{*} = \underset{\textbf{q}}{\operatorname{argmax}} P(\textbf{o,q}| \Lambda)
\end{equation}
We also follow this simplification often employed by methods using HMM models. Thus we assume $P(\textbf{o}|\Lambda_1) > P(\textbf{o}|\Lambda_2)$ if $P(\textbf{o}, \textbf{q}^{*}|\Lambda_1) > P(\textbf{o}, \textbf{q}^{*}|\Lambda_2)$. (This would be referred to as  assumption \textbf{A1}). 
%For the ML estimate, we need to find $\underset{\Lambda}{\operatorname{argmax}}  P(\textbf{o}|\Lambda)$. As in most methods for learning HMMs, we assume that $\underset{\Lambda}{\operatorname{argmax}}  P(\textbf{o}|\Lambda) = \underset{\Lambda}{\operatorname{argmax}} P(\textbf{o},\textbf{q}^*_{\Lambda}| \Lambda)$. 

%Like in \cite{laxman_hmm}, we also make the following assumption
%\begin{equation*}
%\textbf{A1 : }\underset{\Lambda}{\operatorname{argmax}}  P(\textbf{o}|\Lambda) = \underset{\Lambda}{\operatorname{argmax}} P(\textbf{o},\textbf{q}^*_{\Lambda}| \Lambda),
%\end{equation*}
%where the maximum is over the set of HMMs of interest. 

%In our case, the symbol set is the set of event-types and the observed output sequence is the event sequence. For our specialized class of HMMs, we can associate every pair of episodes with a unique HMM of this class. 
Let $\Lambda_{\alpha \beta}$ denote the model corresponding to the pair of episodes, $\alpha$ and $\beta$. We give full description of this model below. For the sake of simplicity, we consider that both are $N$-node episodes. The model depends on whether or not the two episodes share any event types and hence we consider two separate cases (wherever necessary):
\begin{itemize}
	\item \textbf{Case-I: }$\alpha$ and $\beta$ have no common event-types, i.e $\alpha[i]\neq \beta[j]$ $ \forall i,j \in \{ 1,2,...,N \}$.
	\item \textbf{Case-II: }$\alpha$ and $\beta$ have some common event-types. 
\end{itemize}
\subsubsection*{\textbf{The State Space}}
The number of states in the HMM is $4N^2+1$. The state space can be partitioned into two parts: episode states, $\mathcal{S}_e$, comprising of $2N^2$ states and noise states, $\mathcal{S}_n$, comprising of $2N^2+1$ states. Episode states are denoted by $S^k_{i,j}$, $k\in \{ 1,2 \}, \; i,j \in \{ 1,2,..,N \}$. The noise states are given by $N^k_{i,j}$, $k\in \{ 1,2 \}, \; i,j \in \{ 1,2,..,N \}$, and the state $N^0_{1,1}$. 
\subsubsection*{\textbf{Emission structure}}
The symbol probability distribution for the episode states is a delta function. The episode state $S^1_{i,j}$ emits the symbol $\alpha[i]$ with probability 1, whereas $S^2_{i,j}$ emits the symbol $\beta[j]$ with probability 1. For each noise state, the symbol probability distribution is  uniform over the alphabet set $\mathcal{E}$. (We denote $|\mathcal{E}|=M$). 
\subsubsection*{\textbf{Transition structure}}
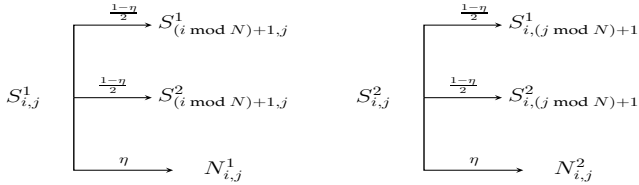
\begin{figure}[h]
	\centering
	\resizebox{9cm}{2.5cm}{
		\begin{tikzpicture}[node distance=2cm]
		\node (root) [box] {\Large $S^1_{i,j}$};
		\node (child2) [box, right of=root, xshift=2cm] { \Large $S^2_{(i\text{ mod }N) +1,j}$};
		\node (child1) [box, above of=child2] {\Large $S^1_{(i\text{ mod }N) +1,j}$};
		\node (child3) [box, below of=child2] {\Large $N^1_{i,j}$};
		
		\draw [arrow] (root.east) |- node[above, xshift=1cm]{$\frac{1-\eta}{2}$}(child1.west);
		\draw [arrow] (root.east) -- node[above]{$\frac{1-\eta}{2}$}(child2.west);
		\draw [arrow] (root.east) |- node[above, xshift=1cm]{$\eta$}(child3.west);
		
		\node (root1) [box, right of=child2, xshift=1cm] {\Large $S^2_{i,j}$};
		\node (child12) [box, right of=root1, xshift=2cm] {\Large $S^2_{i,(j\text{ mod }N) +1}$};
		\node (child11) [box, above of=child12] {\Large $S^1_{i,(j\text{ mod }N) +1}$};
		\node (child13) [box, below of=child12] {\Large $N^2_{i,j}$};
		
		\draw [arrow] (root1.east) |- node[above, xshift=1cm]{$\frac{1-\eta}{2}$}(child11.west);
		\draw [arrow] (root1.east) -- node[above]{$\frac{1-\eta}{2}$}(child12.west);
		\draw [arrow] (root1.east) |- node[above, xshift=1cm]{$\eta$}(child13.west);
		\end{tikzpicture}}
	\caption{Episode state Transition Structure}
	\label{fig:transition}
\end{figure}

\begin{figure}[h]
	\centering
	\captionsetup{width=0.6\textwidth}
	\resizebox{9cm}{2.5cm}{
		\begin{tikzpicture}[node distance=2cm]
		\node (root) [box] {\Large $S^1_{i,(j\text{ mod }N)+1}$};
		\node (child2) [box, right of=root, xshift=2cm, yshift=-1.5cm] {\Large $N^1_{i,(j\text{ mod }N)+1}$};
		\node (child1) [box, above of=child2, xshift=1cm,yshift=1.5cm] {\Large $S^1_{(i\text{ mod }N) +1,((j+1)\text{ mod }N)+1}$};

		\draw [arrow] (root.east) |- node[above, xshift=0.2cm]{$1-\eta$}(child1.west);
		\draw [arrow] (root.east) |- node[above, xshift=0.7cm]{$\eta$}(child2.west);

		\node (root1) [box,right of=root ,xshift=5cm] {\Large $S^2_{(i\text{ mod }N)+1,j}$};
		\node (child12) [box, right of=root1, xshift=2cm, yshift=-1.5cm] {\Large $ N^2_{(i\text{ mod }N)+1,j}$};
		\node (child11) [box, above of=child12, xshift=1cm, yshift=1.5cm] {\Large $S^2_{((i+1)\text{ mod }N)+1,(j\text{ mod }N)+1}$};

		\draw [arrow] (root1.east) |- node[above, xshift=0.2cm]{$1-\eta$}(child11.west);
		\draw [arrow] (root1.east) |- node[above, xshift=0.7cm]{$\eta$}(child12.west);
		\end{tikzpicture}}
	\caption{Transition Structure when $\alpha[(i\text{ mod }N)+1]=\beta[(j\text{ mod }N)+1]$}
	\label{fig:transition1}
\end{figure}
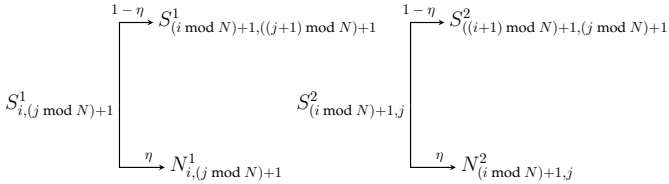

%The number of states in the HMM is $4N^2+1$. The state space can be partitioned into two parts: episode states, $\mathcal{S}_e$, comprising of $2N^2$ states and noise states, $\mathcal{S}_n$, comprising of $2N^2+1$ states. Episode states are denoted by $S^k_{i,j}$, $k\in \{ 1,2 \}, \; i,j \in \{ 1,2,..,N \}$. The noise states are given by $N^k_{i,j}$ , $k\in \{ 1,2 \}, \; i,j \in \{ 1,2,..,N \}$, and the state $N^0_{1,1}$. 
%$S^k_{i,j}$ can be interpreted as the state emitting $\alpha[i]$ or $\beta[j]$.  \newline
Under \textbf{Case-I} (where $\alpha$ and $\beta$ do not share any event types), 
the state transition probabilities out of episode states are given by Fig.~\ref{fig:transition}. Under \textbf{Case-II} also (where $\alpha$ and $\beta$ may share some event types), the transition probabilities out of episode states are as given by  the state transition structure of Fig \ref{fig:transition} except 
for the states $S^1_{i,(j\text{ mod }N) +1}$ and $S^2_{(i\text{ mod }N)+1,j}$ where $i,j$ are such that
 $\alpha[(i\text{ mod }N)+1]=\beta[(j\text{ mod }N)+1]$. For such $i,j$ the transition probabilities are as given in Fig.~\ref{fig:transition1}. \newline
%Under \textbf{Case-I} (where $\alpha$ and $\beta$ do not share any event types), 
%the state transition probabilities out of episode states are given by Fig.~\ref{fig:transition}. Under \textbf{Case-II} also (where $\alpha$ and $\beta$ may share some event types), the transition probabilities out of episode states are as given by  the state transition structure of Fig \ref{fig:transition} except 
%for the states $S^1_{i,(j\text{ mod }N) +1}$ and $S^2_{(i\text{ mod }N)+1,j}$ where $i,j$ are such that
 %$\alpha[(i\text{ mod }N)+1]=\beta[(j\text{ mod }N)+1]$. For such $i,j$ the transition probabilities are as given in Fig.~\ref{fig:transition1}. \newline

\begin{figure}[h]
	\centering
	\resizebox{9cm}{2.5cm}{
		\begin{tikzpicture}[node distance=2cm]
		\node (root) [box] {\Large $N^1_{i,j}$};
		\node (child2) [box, right of=root, xshift=2cm] {\Large $S^2_{(i\text{ mod }N) +1,j}$};
		\node (child1) [box, above of=child2] {\Large $S^1_{(i\text{ mod }N) +1,j}$};
		\node (child3) [box, below of=child2] {\Large $N^1_{i,j}$};
		
		\draw [arrow] (root.east) |- node[above, xshift=1cm]{$\frac{1-\eta}{2}$}(child1.west);
		\draw [arrow] (root.east) -- node[above]{$\frac{1-\eta}{2}$}(child2.west);
		\draw [arrow] (root.east) |- node[above, xshift=1cm]{$\eta$}(child3.west);
		
		\node (root1) [box, right of=child2, xshift=1cm] {\Large $N^2_{i,j}$};
		\node (child12) [box, right of=root1, xshift=2cm] {\Large $S^2_{i,(j\text{ mod }N) +1}$};
		\node (child11) [box, above of=child12] {\Large $S^1_{i,(j\text{ mod }N) +1}$};
		\node (child13) [box, below of=child12] {\Large $N^2_{i,j}$};
		
		\draw [arrow] (root1.east) |- node[above, xshift=1cm]{$\frac{1-\eta}{2}$}(child11.west);
		\draw [arrow] (root1.east) -- node[above]{$\frac{1-\eta}{2}$}(child12.west);
		\draw [arrow] (root1.east) |- node[above, xshift=1cm]{$\eta$}(child13.west);
		\end{tikzpicture}}
	\caption{Noise state Transition Structure}
	\label{fig:noise_transition}
\end{figure}
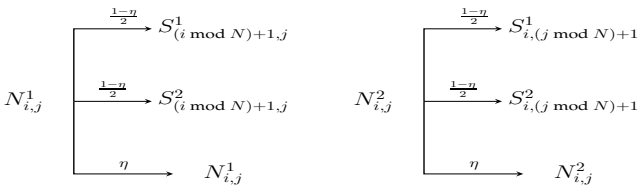

For all the noise states, $N^k_{i,j}$, $k\in \{ 1,2 \}$ the transition structure is as shown in Fig.~\ref{fig:noise_transition}. The noise state $N^0_{1,1}$,  can transit with $\frac{1-\eta}{2}$ probability to each of the episode states $S^2_{1,1}$ and $S^1_{1,1}$ or remain in $N^0_{1,1}$  with $\eta$ probability. 

%The noise state $N^0_{1,1}$,  can transit with $\frac{1-\eta}{2}$ probability to each of the episode states $S^2_{1,1}$ and $S^1_{1,1}$ or remain in $N^0_{1,1}$  with $\eta$ probability.

It may be noted that all transition probabilities are determined by a single parameter, $\eta$, which is called the noise parameter. The values of individual transition probabilities are fixed in an intuitively simple manner. From any state, transitions into a noise state has probability $\eta$. The remaining probability is equally divided between all reachable episode states.

One can intuitively see the logic of the state transition structure also. Recall that in state $S^1_{i,j}$ we emit symbol $\alpha[i]$. So, after this we can either go to $S^1_{(i+1),j}$ to emit the next event type of $\alpha$ or go to $S^2_{(i+1),j}$ to now emit an event type from $\beta$. This allows for arbitrary overlap of occurrences of $\alpha$ and $\beta$. Similarly from $S^2_{i,j}$ (after emitting $\beta[j]$) we can either go to $S^2_{i,(j+1)}$ or $S^1_{(i,(j+1)}$. Since event types constituting occurrence of an episode need not be contiguous, from the episode states we can go to the noise states and cycle there zero or more times before coming back to episode states. After emitting the last event types of, say, $\alpha$, the next event type of $\alpha$ that can be emitted is $\alpha[1]$. Hence, from $S^1_{N,j}$ we should go to either $S^1_{1,j}$ or $S^2_{1,j}$ (or a noise state). That is why in the transition structure as given, whenever an index is incremented it is always with respect to modulo $N$. 

All the above is fine when $\alpha$ and $\beta$ do not share event types. Suppose they share an event type. When that event type appears in the data it could be part of an occurrence of only $\alpha$ or that of only $\beta$ or neither. These possibilities are all accounted for by the above transition structure. However, there is one more possibility, namely, it is part of an occurrence of both $\alpha$ as well as $\beta$; that is, the two occurrences share an event. The transition structure given in Fig.~\ref{fig:transition1} ensures that our generative model includes this possibility too. 

%\subsubsection*{\textbf{Emission structure}}
%The symbol probability distribution for the episode states is a delta function. The episode state $S^1_{i,j}$ emits the symbol $\alpha[i]$ with probability 1, whereas $S^2_{i,j}$ emits the symbol $\beta[j]$ with probability 1. For each noise state, the symbol probability distribution is  uniform over the alphabet set $\mathcal{E}$. (We denote $|\mathcal{E}|=M$). 
\subsubsection*{\textbf{Initial states}}
If $\alpha[1] \neq \beta[1]$, the initial state  is $N^0_{1,1}$ with probability $\eta$, $S^1_{1,1}$ with probability $\frac{1-\eta}{2}$ and $S^2_{1,1}$ with probability $\frac{1-\eta}{2}$ . If $\alpha[1] = \beta[1]$, the initial state is $N^0_{1,1}$ with probability $\eta$, $S^1_{1,2}$ with probability $1-\eta$.

\subsubsection*{\textbf{An Example}}
\begin{figure}[ht]
	\centering
	\resizebox{8cm}{2cm}{
		\begin{tikzpicture}[node distance=2cm]
		\node (node01) [state] {\Large $\color{noise_color} \bm{N^0_{1,1}} $};
		\node (node02) [state, right of=node01] {\Large$\color{state1_color} \bm{S^1_{1,1}}$};
		\node (node03) [state, right of=node02] {\Large $\color{state2_color} \bm{S^2_{2,1}}$};
		\node (node04) [state, right of=node03] {\Large $\color{noise_color} \bm{N^2_{2,1}}$};
		\node (node05) [state, right of=node04] {\Large $\color{state1_color} \bm{S^1_{2,2}}$};
		\node (node06) [state, right of=node05] {\Large $\color{noise_color} \bm{N^1_{2,2}}$};
		\node (node07) [state, right of=node06] {\Large $\color{state2_color} \bm{S^2_{3,2}}$};
		\node (node08) [state, right of=node07] {\Large $\color{state2_color} \bm{S^2_{3,3}}$};
		\node (node09) [state, right of=node08] {\Large $\color{noise_color} \bm{N^2_{3,3}}$};
		\node (node10) [state, right of=node09] {\Large $\color{state1_color} \bm{S^1_{3,1}}$};
		
		\node (emit01) [io, below of=node01] {$\textbf{\textcolor{orange}{\textit{\Large B}}}$};
		\node (emit02) [io, below of=node02] {$\textbf{\textcolor{orange}{\textit{\Large A}}}$};
		\node (emit03) [io, below of=node03] {$\textbf{\textcolor{orange}{\textit{\Large D}}}$};
		\node (emit04) [io, below of=node04] {$\textbf{\textcolor{orange}{\textit{\Large F}}}$};
		\node (emit05) [io, below of=node05] {$\textbf{\textcolor{orange}{\textit{\Large B}}}$};
		\node (emit06) [io, below of=node06] {$\textbf{\textcolor{orange}{\textit{\Large G}}}$};
		\node (emit07) [io, below of=node07] {$\textbf{\textcolor{orange}{\textit{\Large B}}}$};
		\node (emit08) [io, below of=node08] {$\textbf{\textcolor{orange}{\textit{\Large E}}}$};
		\node (emit09) [io, below of=node09] {$\textbf{\textcolor{orange}{\textit{\Large C}}}$};
		\node (emit10) [io, below of=node10] {$\textbf{\textcolor{orange}{\textit{\Large C}}}$};

		\draw [arrow] (node01.east) -- (node02.west);
		\draw [arrow] (node02.east) -- (node03.west);
		\draw [arrow] (node03.east) -- (node04.west);
		\draw [arrow] (node04.east) -- (node05.west);
		\draw [arrow] (node05.east) -- (node06.west);
		\draw [arrow] (node06.east) -- (node07.west);
		\draw [arrow] (node07.east) -- (node08.west);
		\draw [arrow] (node08.east) -- (node09.west);
		\draw [arrow] (node09.east) -- (node10.west);
		
		\draw [arrow] (node01.south) -- (emit01.north);
		\draw [arrow] (node02.south) -- (emit02.north);
		\draw [arrow] (node03.south) -- (emit03.north);
		\draw [arrow] (node04.south) -- (emit04.north);
		\draw [arrow] (node05.south) -- (emit05.north);
		\draw [arrow] (node06.south) -- (emit06.north);
		\draw [arrow] (node07.south) -- (emit07.north);
		\draw [arrow] (node08.south) -- (emit08.north);
		\draw [arrow] (node09.south) -- (emit09.north);
		\draw [arrow] (node10.south) -- (emit10.north);
		\end{tikzpicture}}
	%	\caption{Episode state Transition Structure}
	%	\label{fig:transition}
	\resizebox{8cm}{2cm}{
		\begin{tikzpicture}[node distance=2cm]
		\node (node01) [state] {\Large $\color{state2_color} \bm{S^2_{1,1}}$};
		\node (node02) [state, right of=node01] {\Large $\color{state2_color} \bm{S^2_{1,2}}$};
		\node (node03) [state, right of=node02] {\Large $\color{noise_color}\bm{N^2_{1,2}}$};
		\node (node04) [state, right of=node03] {\Large $\color{state1_color} \bm{S^1_{1,3}}$};
		\node (node05) [state, right of=node04] {\Large $\color{state2_color} \bm{S^2_{2,3}}$};
		\node (node06) [state, right of=node05] {\Large $\color{noise_color} \bm{N^2_{2,3}}$};
		\node (node07) [state, right of=node06] {\Large $\color{state1_color} \bm{S^1_{2,1}}$};
		\node (node08) [state, right of=node07] {\Large $\color{noise_color} \bm{N^1_{2,1}}$};
		\node (node09) [state, right of=node08] {\Large $\color{noise_color} \bm{N^1_{2,1}}$};
		\node (node10) [state, right of=node09] {\Large $\color{state1_color} \bm{S^1_{3,1}}$};
		
		\node (emit01) [io, below of=node01] {$\textbf{\textcolor{orange}{\textit{\Large D}}}$};
		\node (emit02) [io, below of=node02] {$\textbf{\textcolor{orange}{\textit{\Large B}}}$};
		\node (emit03) [io, below of=node03] {$\textbf{\textcolor{orange}{\textit{\Large G}}}$};
		\node (emit04) [io, below of=node04] {$\textbf{\textcolor{orange}{\textit{\Large A}}}$};
		\node (emit05) [io, below of=node05] {$\textbf{\textcolor{orange}{\textit{\Large E}}}$};
		\node (emit06) [io, below of=node06] {$\textbf{\textcolor{orange}{\textit{\Large B}}}$};
		\node (emit07) [io, below of=node07] {$\textbf{\textcolor{orange}{\textit{\Large B}}}$};
		\node (emit08) [io, below of=node08] {$\textbf{\textcolor{orange}{\textit{\Large F}}}$};
		\node (emit09) [io, below of=node09] {$\textbf{\textcolor{orange}{\textit{\Large A}}}$};
		\node (emit10) [io, below of=node10] {$\textbf{\textcolor{orange}{\textit{\Large C}}}$};

		\draw [arrow] (node01.east) -- (node02.west);
		\draw [arrow] (node02.east) -- (node03.west);
		\draw [arrow] (node03.east) -- (node04.west);
		\draw [arrow] (node04.east) -- (node05.west);
		\draw [arrow] (node05.east) -- (node06.west);
		\draw [arrow] (node06.east) -- (node07.west);
		\draw [arrow] (node07.east) -- (node08.west);
		\draw [arrow] (node08.east) -- (node09.west);
		\draw [arrow] (node09.east) -- (node10.west);
		
		\draw [arrow] (node01.south) -- (emit01.north);
		\draw [arrow] (node02.south) -- (emit02.north);
		\draw [arrow] (node03.south) -- (emit03.north);
		\draw [arrow] (node04.south) -- (emit04.north);
		\draw [arrow] (node05.south) -- (emit05.north);
		\draw [arrow] (node06.south) -- (emit06.north);
		\draw [arrow] (node07.south) -- (emit07.north);
		\draw [arrow] (node08.south) -- (emit08.north);
		\draw [arrow] (node09.south) -- (emit09.north);
		\draw [arrow] (node10.south) -- (emit10.north);
		\end{tikzpicture}}
	\resizebox{8cm}{2cm}{
		\begin{tikzpicture}[node distance=2cm]
		\node (node01) [state] {\Large $\color{noise_color} \bm{N^0_{1,1}}$};
		\node (node02) [state, right of=node01] {\Large $\color{state1_color} \bm{S^1_{1,1}}$};
		\node (node03) [state, right of=node02] {\Large $\color{noise_color} \bm{N^1_{1,1}}$};
		\node (node04) [state, right of=node03] {\Large $\color{state2_color} \bm{S^2_{2,1}}$};
		\node (node05) [state, right of=node04] {\Large $\color{state2_color} \bm{S^2_{3,2}}$};
		\node (node06) [state, right of=node05] {\Large $\color{noise_color} \bm{N^2_{3,2}}$};
		\node (node07) [state, right of=node06] {\Large $\color{state1_color} \bm{S^1_{3,3}}$};
		\node (node08) [state, right of=node07] {\Large $\color{noise_color} \bm{N^1_{3,3}}$};
		\node (node09) [state, right of=node08] {\Large $\color{state2_color} \bm{S^2_{1,3}}$};
		\node (node10) [state, right of=node09] {\Large $\color{noise_color} \bm{N^2_{1,3}}$};
		
		\node (emit01) [io, below of=node01] {$\textbf{\textcolor{orange}{\textit{\Large F}}}$};
		\node (emit02) [io, below of=node02] {$\textbf{\textcolor{orange}{\textit{\Large A}}}$};
		\node (emit03) [io, below of=node03] {$\textbf{\textcolor{orange}{\textit{\Large C}}}$};
		\node (emit04) [io, below of=node04] {$\textbf{\textcolor{orange}{\textit{\Large D}}}$};
		\node (emit05) [io, below of=node05] {$\textbf{\textcolor{orange}{\textit{\Large B}}}$};
		\node (emit06) [io, below of=node06] {$\textbf{\textcolor{orange}{\textit{\Large B}}}$};
		\node (emit07) [io, below of=node07] {$\textbf{\textcolor{orange}{\textit{\Large C}}}$};
		\node (emit08) [io, below of=node08] {$\textbf{\textcolor{orange}{\textit{\Large G}}}$};
		\node (emit09) [io, below of=node09] {$\textbf{\textcolor{orange}{\textit{\Large E}}}$};
		\node (emit10) [io, below of=node10] {$\textbf{\textcolor{orange}{\textit{\Large F}}}$};

		\draw [arrow] (node01.east) -- (node02.west);
		\draw [arrow] (node02.east) -- (node03.west);
		\draw [arrow] (node03.east) -- (node04.west);
		\draw [arrow] (node04.east) -- (node05.west);
		\draw [arrow] (node05.east) -- (node06.west);
		\draw [arrow] (node06.east) -- (node07.west);
		\draw [arrow] (node07.east) -- (node08.west);
		\draw [arrow] (node08.east) -- (node09.west);
		\draw [arrow] (node09.east) -- (node10.west);
		
		\draw [arrow] (node01.south) -- (emit01.north);
		\draw [arrow] (node02.south) -- (emit02.north);
		\draw [arrow] (node03.south) -- (emit03.north);
		\draw [arrow] (node04.south) -- (emit04.north);
		\draw [arrow] (node05.south) -- (emit05.north);
		\draw [arrow] (node06.south) -- (emit06.north);
		\draw [arrow] (node07.south) -- (emit07.north);
		\draw [arrow] (node08.south) -- (emit08.north);
		\draw [arrow] (node09.south) -- (emit09.north);
		\draw [arrow] (node10.south) -- (emit10.north);
		\end{tikzpicture}}
	
	\resizebox{8cm}{2cm}{
		\begin{tikzpicture}[node distance=2cm]
		\node (node01) [state] {\Large $\color{state2_color} \bm{S^2_{1,1}}$};
		\node (node02) [state, right of=node01] {\Large $\color{noise_color} \bm{N^2_{1,1}}$};
		\node (node03) [state, right of=node02] {\Large $\color{noise_color} \bm{N^2_{1,1}}$};
		\node (node04) [state, right of=node03] {\Large $\color{state1_color} \bm{S^1_{1,2}}$};
		\node (node05) [state, right of=node04] {\Large $\color{noise_color} \bm{N^1_{1,2}}$};
		\node (node06) [state, right of=node05] {\Large $\color{state1_color} \bm{S^1_{2,2}}$};
		\node (node07) [state, right of=node06] {\Large $\color{state2_color} \bm{S^2_{3,2}}$};
		\node (node08) [state, right of=node07] {\Large $\color{state2_color} \bm{S^2_{3,3}}$};
		\node (node09) [state, right of=node08] {\Large $\color{state1_color} \bm{S^1_{3,1}}$};
		\node (node10) [state, right of=node09] {\Large $\color{noise_color} \bm{N^1_{3,1}}$};
		
		\node (emit01) [io, below of=node01] {$\textbf{\textcolor{orange}{\textit{\Large D}}}$};
		\node (emit02) [io, below of=node02] {$\textbf{\textcolor{orange}{\textit{\Large G}}}$};
		\node (emit03) [io, below of=node03] {$\textbf{\textcolor{orange}{\textit{\Large E}}}$};
		\node (emit04) [io, below of=node04] {$\textbf{\textcolor{orange}{\textit{\Large A}}}$};
		\node (emit05) [io, below of=node05] {$\textbf{\textcolor{orange}{\textit{\Large B}}}$};
		\node (emit06) [io, below of=node06] {$\textbf{\textcolor{orange}{\textit{\Large B}}}$};
		\node (emit07) [io, below of=node07] {$\textbf{\textcolor{orange}{\textit{\Large B}}}$};
		\node (emit08) [io, below of=node08] {$\textbf{\textcolor{orange}{\textit{\Large E}}}$};
		\node (emit09) [io, below of=node09] {$\textbf{\textcolor{orange}{\textit{\Large C}}}$};
		\node (emit10) [io, below of=node10] {$\textbf{\textcolor{orange}{\textit{\Large G}}}$};

		\draw [arrow] (node01.east) -- (node02.west);
		\draw [arrow] (node02.east) -- (node03.west);
		\draw [arrow] (node03.east) -- (node04.west);
		\draw [arrow] (node04.east) -- (node05.west);
		\draw [arrow] (node05.east) -- (node06.west);
		\draw [arrow] (node06.east) -- (node07.west);
		\draw [arrow] (node07.east) -- (node08.west);
		\draw [arrow] (node08.east) -- (node09.west);
		\draw [arrow] (node09.east) -- (node10.west);
		
		\draw [arrow] (node01.south) -- (emit01.north);
		\draw [arrow] (node02.south) -- (emit02.north);
		\draw [arrow] (node03.south) -- (emit03.north);
		\draw [arrow] (node04.south) -- (emit04.north);
		\draw [arrow] (node05.south) -- (emit05.north);
		\draw [arrow] (node06.south) -- (emit06.north);
		\draw [arrow] (node07.south) -- (emit07.north);
		\draw [arrow] (node08.south) -- (emit08.north);
		\draw [arrow] (node09.south) -- (emit09.north);
		\draw [arrow] (node10.south) -- (emit10.north);
		\end{tikzpicture}}
	\caption{First 10 events of four sample output sequences}
\label{fig:example_output}
\end{figure}
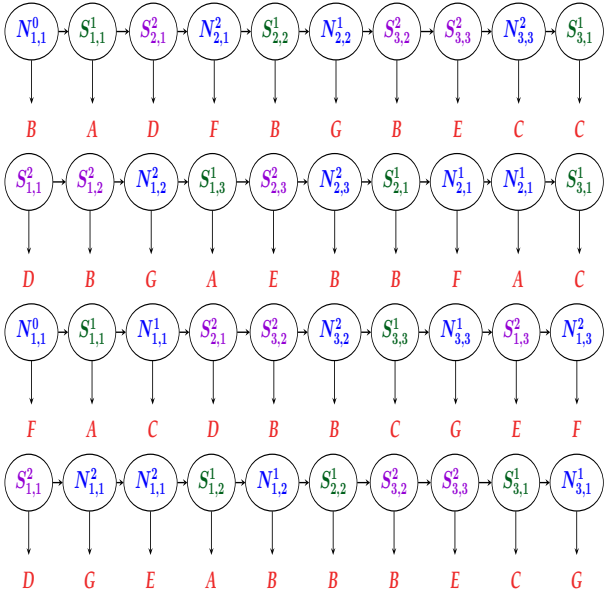

Consider a model $\Lambda_{\alpha \beta}$, where $\alpha=A\to B\to C$ and $\beta = D\to B\to E$. Let the alphabet set $\mathcal{E}=\{ A,B,C,D,E,F,G \}$. We show a few example state sequences and output sequences of length 10 that can be emitted by $\Lambda_{\alpha \beta}$ in Fig.~\ref{fig:example_output}. As can be seen from the figure, the output sequence contains occurrences of $\alpha$ and $\beta$ that may be arbitrarily interleaved.  Here we have $\alpha[2] = \beta[2]$. Hence transitions out of episode states $S^1_{12}, S^2_{21}$ are as given in Fig.~\ref{fig:transition1} and for all other episode states they are as given in Fig.~\ref{fig:transition}. The special transition structure for $S^1_{12}, S^2_{21}$ allows some occurrences of $\alpha$ and $\beta$ to share an event (of event type $B$) as can be seen in row-3 (the transition from $S^2_{21}$ to $S^2_{32}$)  of Fig.~\ref{fig:example_output}.

\subsection{Analysis}
In this section we derive expressions for the likelihood (of a joint state and output sequence) of our HMM model and use this to compare likelihoods of models corresponding to different pairs of episodes. The expressions depend on whether or not the pairs of episodes share event types and hence the two cases are dealt with separately. 

\noindent
{\bf In all our analysis we assume $\eta < \frac{M}{M+8}$ where $M=|\mathcal{E}|$}.
\subsubsection{Case-I}
Here, $\alpha[i] \neq \beta[j] \quad \forall i,j \in \{ 1,2,\dots,N \}$. Hence, all episode states have only $\frac{1-\eta}{2}$ transition into them. Decomposing any state sequence into two sub-sequences $\textbf{q}_e$ and $\textbf{q}_n$, corresponding to the episode and noise states, we have the following observation: in equation \eqref{eq:5}, whenever the transition probability $p_{q_{t-1}q_{t}}$ is $(1-\eta)/2$, the state $q_t$ has to be an episode state, and hence the  $b_{q_t}(o_t)$ is either $1$ or $0$. Similarly, whenever $p_{q_{t-1}q_{t}}$ is $\eta$, the corresponding $b_{q_t}(o_t)$ is $\frac{1}{M}$. Thus for any state sequence with non-zero probability, we can write the joint probability as 
\begin{equation}
\label{eq:7}
P(\textbf{o,q}| \Lambda) = \Bigg(\frac{\eta}{M} \Bigg)^{|\textbf{q}_n|} \Bigg( \frac{1-\eta}{2} \Bigg)^{|\textbf{q}_e|}
\end{equation}

Here, $|\textbf{q}_n|$ and $|\textbf{q}_e|$ denote lengths of the respective sub-sequences. Since $|\textbf{q}_e| + |\textbf{q}_n|=|\textbf{q}|=T$ (the length of output or event sequence), \eqref{eq:7} can be written as 
\begin{equation}
P(\textbf{o,q}| \Lambda) = \Bigg(\frac{\eta}{M} \Bigg)^{T} \Bigg( \frac{(1-\eta)M}{2 \eta} \Bigg)^{|\textbf{q}_e|}
\end{equation}
Under our assumption we have $\eta < \frac{M}{M+8}$, and hence $ \frac{(1-\eta)M}{2 \eta}  > 1$. Then, $P(\textbf{o,q}| \Lambda)$ is monotonically increasing with $|\textbf{q}_e|$. Thus, the most likely state sequence is the one that spends the longest time in episode states. Due to constraints imposed on the state transition structure, in any state sequence of $\Lambda_{\alpha \beta}$ having non-zero probability, the episode states corresponding to a particular episode have to occur in sequence. Moreover, when a particular episode state $S^1_{i,j}$ or $S^2_{i,j}$ is revisited, it implies one cycle of all the event-types corresponding to that episode have been emitted. Suppose $f^{*}_\alpha$ and $f^{*}_\beta$ are the maximum possible number of non-overlapping occurrences of $\alpha$ and $\beta$ respectively in $\textbf{o}$. Since, $\alpha$ and $\beta$ do not share any event type and at each episode state we emit one symbol, the most likely state sequence has at least $N(f^{*}_\alpha+f^{*}_\beta)$ number of episode states in it, i.e $|\textbf{q}^*_{e}| \geq Nf^{*}_{\alpha} + Nf^{*}_{\beta}$.

For the sake of simplicity, we make the assumption (referred to as  \textbf{A2}) that there is no state sequence with non-zero probability that includes any incomplete occurrence of either of the episodes.\footnote{ 
%In other words, in any state sequence with non-zero probability the last episode state is either $S^1_{N,1}$ or $S^2_{1,N}$. 
We can ensure that the assumption \textbf{A2} always holds by modifying our model by adding an extra symbol  (``end-of-sequence" marker) at the end of the output sequence \textbf{o} and modifying the symbol probability distribution of the noise states $N^1_{N,1}$ and $N^2_{1,N}$ by following the trick used with HMMs for single episodes as in \cite{laxman_ms}.}

Under the assumption \textbf{A2}, we have $|\textbf{q}^*_{e}| = Nf^{*}_{\alpha} + Nf^{*}_{\beta}$. 
%Under the consideration that $\eta < \frac{M}{M+2}$, we have $\frac{(1-\eta)M}{2\eta}>1$. 
Consider two models $\Lambda_{\alpha \beta}$ and $\Lambda_{\alpha \gamma}$. 
\begin{align*}
P(\textbf{o},\textbf{q}^*| \Lambda_{\alpha \beta}) &= \Bigg(\frac{\eta}{M} \Bigg)^{T} \Bigg( \frac{(1-\eta)M}{2 \eta} \Bigg)^{Nf^{*}_{\alpha} + Nf^{*}_{\beta} } \\
P(\textbf{o},\textbf{q}^*| \Lambda_{\alpha \gamma}) &= \Bigg(\frac{\eta}{M} \Bigg)^{T} \Bigg( \frac{(1-\eta)M}{2 \eta} \Bigg)^{Nf^{*}_{\alpha} + Nf^{*}_{\gamma} } \\
\implies \frac{P(\textbf{o},\textbf{q}^*| \Lambda_{\alpha \beta})}{P(\textbf{o},\textbf{q}^*| \Lambda_{\alpha \gamma})} &=  \Bigg( \frac{(1-\eta)M}{2 \eta} \Bigg)^{N(f^{*}_{\beta} - f^{*}_{\gamma}) }
\end{align*}
Hence, under assumption \textbf{A1}, 
%\begin{equation*}
%\frac{P(\textbf{o}| \Lambda_{\alpha \beta})}{P(\textbf{o}| \Lambda_{\alpha \gamma})} =  \Bigg( \frac{(1-\eta)M}{2 \eta} \Bigg)^{N(f^{*}_{\beta} - f^{*}_{\gamma}) }
%\end{equation*}
if $f^{*}_\beta > f^{*}_\gamma$, we have $P(\textbf{o}| \Lambda_{\alpha \beta}) > P(\textbf{o}| \Lambda_{\alpha \gamma})$. 
%So, data likelihood is more for the more frequent one. 
This essentially implies that given an already selected episode (the episode $\alpha$ here), if we want to select the next one from the set of episodes that do not share any event type with the already selected one, we should choose the most frequent one from that set.
\subsubsection{Case-II}
In this case, we have some episode states with a $(1-\eta)/2$ transition into them while some have $(1-\eta)$ transition into them. 
%In this case also, we consider that \textbf{A2} holds. 
It should be noted that because of the transition structure, a symbol emitted from an episode state with $1-\eta$ transition into it is part of an occurrence of both the episodes. It means that the event is shared across occurrences of the two episodes. On the other hand, a symbol emitted from an episode state with $(1-\eta)/2$ transition into it is part of an occurrence of only one episode and hence is not shared. Now, we further decompose the episode states part of any state sequence, $\textbf{q}_e$ into two parts $\textbf{q}_1, \textbf{q}_2$. The episode states corresponding to event types that are not shared form $\textbf{q}_1$ while those corresponding to shared ones form $\textbf{q}_2$. Since, every state emits one symbol, we have $|\textbf{q}_n| + |\textbf{q}_1|+ |\textbf{q}_2| = T$.
%Under our assumption that $\eta < \frac{M}{M+4}$ we have 
%We consider the value of $\eta$ to be such that emitting two symbols from non-sharing episode states is more likely than emitting one symbol from the shared episode state and the other from noise state, i.e 
%\begin{align*}
%\[\frac{1-\eta}{2} \times \frac{1-\eta}{2} > (1-\eta) \times \frac{\eta}{M} \]
%\implies \eta &< \frac{M}{M+4}
%\end{align*}
Now, the joint probability  of an output and state sequence is given by 
\begin{align*}
P(\textbf{o,q}| \Lambda) &= \Bigg(\frac{\eta}{M} \Bigg)^{|\textbf{q}_n|} \Bigg( \frac{(1-\eta)}{2} \Bigg)^{|\textbf{q}_1|} \Bigg( 1-\eta \Bigg)^{|\textbf{q}_2|} \\
& = \Bigg(\frac{\eta}{M} \Bigg)^{T} \Bigg( \frac{(1-\eta)M}{2 \eta} \Bigg)^{|\textbf{q}_1|} \Bigg( \frac{(1-\eta)M}{ \eta} \Bigg)^{|\textbf{q}_2|}                      
\end{align*}

%Since, every state emits one symbol, we have $|\textbf{q}_n| + |\textbf{q}_1|+ |\textbf{q}_2| = T$. 
Let us consider a state sequence $\textbf{q}$ (having non-zero probability) that contains  $f_\alpha$ and $f_\beta$ number of occurrences of the episodes $\alpha$ and $\beta$ respectively. Let the number of events shared between these occurrences be $O_{\alpha \beta}$. Then, the no of events covered by the occurrences of the episodes in the output sequence  is $(Nf_{\alpha} + Nf_{\beta} - O_{\alpha \beta}),$ out of which $(Nf_{\alpha} + Nf_{\beta} - 2O_{\alpha \beta})$ number of events are not shared and $O_{\alpha \beta}$ number of events are shared. Under assumption \textbf{A2}, we have $|\textbf{q}_2|= O_{\alpha \beta}$ and $|\textbf{q}_1|= Nf_{\alpha} + Nf_{\beta} - 2O_{\alpha \beta}$. So, for this state sequence, 
\begin{align*}
P(\textbf{o,q}| \Lambda_{\alpha \beta}) &= \Bigg(\frac{\eta}{M} \Bigg)^{T} \Bigg( \frac{(1-\eta)M}{2 \eta} \Bigg)^{Nf_{\alpha} + Nf_{\beta} - 2O_{\alpha \beta}} \\
& \qquad \qquad \qquad \qquad \quad \Bigg( \frac{(1-\eta)M}{ \eta} \Bigg)^{O_{\alpha \beta}} \\
&= \Bigg(\frac{\eta}{M} \Bigg)^{T} \Bigg( \frac{(1-\eta)M}{2 \eta} \Bigg)^{Nf_{\alpha} + Nf_{\beta} }  \\ 
& \qquad \qquad \qquad \qquad \quad \Bigg( \frac{(1-\eta)M}{ 4\eta} \Bigg)^{-O_{\alpha \beta}} \numberthis \label{eq:9}
\end{align*}
For $\eta < \frac{M}{M+8}$, $\frac{(1-\eta)M}{ 4\eta} > 1$. So, we see that the joint probability is an increasing function of the no of occurrences of the episodes, and for fixed $f_\alpha$ and $f_\beta$, a decreasing function of the number of events shared between the occurrences. 

%So, the joint probability for the most likely state sequence (say, $\textbf{q}^*$) is given as 
%\begin{align*}
%P(\textbf{o},\textbf{q}^*| \Lambda_{\alpha \beta}) &= \underset{f_\alpha,f_\beta}{\operatorname{max}} \quad  \underset{O_{\alpha \beta}}{\operatorname{min}} \quad \Bigg(\frac{\eta}{M} \Bigg)^{T} \Bigg( \frac{(1-\eta)M}{2 \eta} \Bigg)^{Nf_{\alpha} + Nf_{\beta} }  \\ 
%& \qquad \qquad \qquad \qquad  \Bigg( \frac{(1-\eta)M}{ 4\eta} \Bigg)^{-O_{\alpha \beta}}  \numberthis \label{eq:10}
%\end{align*}
Let $f^{*}_\alpha$ and $f^{*}_\beta$ be the maximum possible number of non-overlapped occurrences of $\alpha$ and $\beta$ respectively in $\textbf{o}$. So, the most likely state sequence ($\textbf{q}^*$) is the one which emits all the $f^{*}_\alpha+f^{*}_\beta$ number of occurrences from the episode states and among all such state sequences 
%having non-zero probability of emitting $f^{*}_\alpha+f^{*}_\beta$ number of occurrences, the most likely state sequence 
it is the one which shares minimum number of events between these occurrences. Let $O^{*}_{\alpha \beta}$ be the number of shared events corresponding to $\textbf{q}^*$.  Then, from \eqref{eq:9}, 
\begin{align*}
P(\textbf{o},\textbf{q}^*| \Lambda_{\alpha \beta}) &=  \Bigg(\frac{\eta}{M} \Bigg)^{T} \Bigg( \frac{(1-\eta)M}{2 \eta} \Bigg)^{Nf^{*}_{\alpha} + Nf^{*}_{\beta} }  \\ 
& \qquad \qquad \qquad \qquad  \Bigg( \frac{(1-\eta)M}{ 4\eta} \Bigg)^{-O^{*}_{\alpha \beta}}  \numberthis \label{eq:11}
\end{align*}
We will have a similar expression for the model $\Lambda_{\alpha \gamma}$ and hence
\begin{align*}
%P(\textbf{o},\textbf{q}^*| \Lambda_{\alpha \beta}) &=  \Bigg(\frac{\eta}{M} \Bigg)^{T} \Bigg( \frac{(1-\eta)M}{2 \eta} \Bigg)^{Nf^{*}_{\alpha} + Nf^{*}_{\beta} }  \\ 
%& \qquad \qquad \qquad \qquad  \Bigg( \frac{(1-\eta)M}{ 4\eta} \Bigg)^{-O^{*}_{\alpha \beta}} \\
%P(\textbf{o},\textbf{q}^*| \Lambda_{\alpha \gamma}) &=  \Bigg(\frac{\eta}{M} \Bigg)^{T} \Bigg( \frac{(1-\eta)M}{2 \eta} \Bigg)^{Nf^{*}_{\alpha} + Nf^{*}_{\gamma} }  \\ 
%& \qquad \qquad \qquad \qquad  \Bigg( \frac{(1-\eta)M}{ 4\eta} \Bigg)^{-O^{*}_{\alpha \gamma}}   \\
%\implies 
\frac{P(\textbf{o},\textbf{q}^*| \Lambda_{\alpha \beta})}{P(\textbf{o},\textbf{q}^*| \Lambda_{\alpha \gamma})} &=  \Bigg( \frac{(1-\eta)M}{2 \eta} \Bigg)^{N(f^{*}_{\beta} - f^{*}_{\gamma}) } \\
& \qquad \qquad \qquad \qquad  \Bigg( \frac{(1-\eta)M}{ 4\eta} \Bigg)^{O^{*}_{\alpha \gamma}-O^{*}_{\alpha \beta}} \numberthis \label{eq:12}  
\end{align*}
%Taking \textbf{A1} into consideration, 
%\begin{equation*}
%\frac{P(\textbf{o}| \Lambda_{\alpha \beta})}{P(\textbf{o}| \Lambda_{\alpha \gamma})} =  \Bigg( \frac{(1-\eta)M}{2 \eta} \Bigg)^{N(f^{*}_{\beta} - f^{*}_{\gamma}) } \Bigg( \frac{(1-\eta)M}{ 4\eta} \Bigg)^{O^{*}_{\alpha \gamma}-O^{*}_{\alpha \beta}} \numberthis \label{eq:12}
%\end{equation*}
Thus, under assumption \textbf{A1}, we see that if $f_{\beta}=f_{\gamma}$, likelihood is higher for the pair of episodes that share lesser number of events. In general, the relative likelihood of $\Lambda_{\alpha \beta}$ and $ \Lambda_{\alpha \gamma}$ depends both on the frequencies of $\beta$ and $\gamma$ as well as on the difference in their overlaps with $\alpha$.  %Hence, given a already selected episode, if two episodes have same frequency and hence in turn same \textit{Score} \eqref{eq:2},  the episode that share less events with the already selected episode, i.e the one that has higher \textit{Overlap-Score} \eqref{eq:3} should be selected next. 
%In general, it can be observed that the data likelihood is an interplay between the frequency and number of events shared.
To better understand this, let us define two metrics to rate any other episode with respect to episode $\alpha$.

\begin{eqnarray*}
Overlap\text{-}score_1(\beta,\alpha) &=& Nf^{*}_{\beta} - O^{*}_{\alpha \beta} \\
Overlap\text{-}score_2(\beta,\alpha) &=& Nf^{*}_{\beta} - \frac{1}{2} O^{*}_{\alpha \beta}  \label{eq:9.1}
\end{eqnarray*}

We will show that given an episode $\alpha$, if the values of both metric for an episode $\beta$ are higher than those for an episode $\gamma$, then  $\Lambda_{\alpha,\beta}$ has higher data likelihood compared to $\Lambda_{\alpha,\gamma}$ (under our assumption on $\eta$ and under \textbf{A1} and \textbf{A2}). 

\textbf{Case-a:} $f^{*}_{\beta}>f^{*}_{\gamma}, O^{*}_{\alpha \beta}<O^{*}_{\alpha \gamma}$ \newline
Under assumption \textbf{A1}, from \eqref{eq:12}, it is easily seen that $P(\textbf{o}| \Lambda_{\alpha \beta}) > P(\textbf{o}| \Lambda_{\alpha \gamma})$. Also, it is easy to check that $f^{*}_{\beta}>f^{*}_{\gamma}$ and $-O^{*}_{\alpha \beta}>-O^{*}_{\alpha \gamma}$ imply 
%Hence,  $Nf^{*}_{\beta}-O^{*}_{\alpha \beta}>  + Nf^{*}_{\gamma}-O^{*}_{\alpha \gamma}$ and $Nf^{*}_{\beta}-\frac{1}{2}O^{*}_{\alpha \beta}>  + Nf^{*}_{\gamma}-\frac{1}{2}O^{*}_{\alpha \gamma}$ which implies 
$Overlap\text{-}score_1(\beta,\alpha)>Overlap\text{-}score_1(\gamma,\alpha)$
 and $Overlap\text{-}score_2(\beta,\alpha)>Overlap\text{-}score_2(\gamma,\alpha)$

\textbf{Case-b:} $f^{*}_{\beta}>f^{*}_{\gamma}, O^{*}_{\alpha \beta}>O^{*}_{\alpha \gamma}$, \newline 
In this scenario, depending on the values of overlaps, the two metrics for $\beta$ may be greater or smaller than those of $\gamma$. Hence we consider these two sub-cases.
\newline
\textit{Case-b1: }$Overlap\text{-}score_1(\beta,\alpha)>Overlap\text{-}score_1(\gamma,\alpha)$
and $Overlap\text{-}score_2(\beta,\alpha)>Overlap\text{-}score_2(\gamma,\alpha)$.
\newline
Since, $Overlap\text{-}score_1(\beta,\alpha)>Overlap\text{-}score_1(\gamma,\alpha) \implies Nf^{*}_{\beta} - O^{*}_{\alpha \beta} > Nf^{*}_{\gamma} - O^{*}_{\alpha \gamma} \implies Nf^{*}_{\beta} - Nf^{*}_{\gamma} > O^{*}_{\alpha \beta} -O^{*}_{\alpha \gamma}$, we have from \eqref{eq:12},
\begin{align*}
\frac{P(\textbf{o}, \textbf{q}^*| \Lambda_{\alpha \beta})}{P(\textbf{o}, \textbf{q}^*| \Lambda_{\alpha \gamma})} = \frac{\Bigg( \frac{(1-\eta)M}{2 \eta} \Bigg)^{N(f^{*}_{\beta} - f^{*}_{\gamma}) }}{\Bigg( \frac{(1-\eta)M}{ 4\eta} \Bigg)^{O^{*}_{\alpha \beta}-O^{*}_{\alpha \gamma}}} \qquad \qquad \qquad \qquad \\
= 2^{N(f^{*}_{\beta} - f^{*}_{\gamma})} \Bigg( \frac{(1-\eta)M}{ 4\eta} \Bigg)^{N(f^{*}_{\beta} - f^{*}_{\gamma}) - (O^{*}_{\alpha \beta}-O^{*}_{\alpha \gamma})} > 1 \\ \implies P(\textbf{o}| \Lambda_{\alpha \beta}) > P(\textbf{o}| \Lambda_{\alpha \gamma})  \mbox{~~~(under assumption \textbf{A1})} 
%\qquad \qquad \qquad \qquad
\end{align*}
\textit{Case-b2: }$Overlap\text{-}score_1(\gamma,\alpha)>Overlap\text{-}score_1(\beta,\alpha)$
and $Overlap\text{-}score_2(\gamma,\alpha)>Overlap\text{-}score_2(\beta,\alpha)$
\newline
Since, $Overlap\text{-}score_2(\gamma,\alpha)>Overlap\text{-}score_2(\beta,\alpha) \implies Nf^{*}_{\gamma} -\frac{1}{2} O^{*}_{\alpha \gamma} > Nf^{*}_{\beta} -\frac{1}{2} O^{*}_{\alpha \beta} \implies Nf^{*}_{\beta} - Nf^{*}_{\gamma} < \frac{1}{2}(O^{*}_{\alpha \beta} -O^{*}_{\alpha \gamma})$. Let $Nf^{*}_{\beta} - Nf^{*}_{\gamma}$ be $x$. Then we can write $(O^{*}_{\alpha \beta} -O^{*}_{\alpha \gamma})=2x+ \xi$, where $\xi > 0$. Since we assume $\eta < \frac{M}{M+8}$, we have $\frac{(1-\eta)M}{ 8\eta} > 1$. Now, from \eqref{eq:12},
\begin{align*}
\frac{P(\textbf{o}, \textbf{q}^*| \Lambda_{\alpha \beta})}{P(\textbf{o}, \textbf{q}^*| \Lambda_{\alpha \gamma})} &= \frac{\Bigg( \frac{(1-\eta)M}{2 \eta} \Bigg)^{x}}{\Bigg( \frac{(1-\eta)M}{ 4\eta} \Bigg)^{2x+\xi}}  \\
&= \frac{2^{x}}{\Bigg( \frac{(1-\eta)M}{ 4\eta} \Bigg)^{x+\xi}}  \\ 
&= \frac{1}{\Bigg( \frac{(1-\eta)M}{ 8\eta} \Bigg)^{x} \Bigg( \frac{(1-\eta)M}{ 4\eta} \Bigg)^{\xi}}  < 1\\ 
&\implies P(\textbf{o}| \Lambda_{\alpha \gamma}) > P(\textbf{o}| \Lambda_{\alpha \beta}) \mbox{~~~(under  \textbf{A1})}
%\qquad \qquad \qquad \qquad
\end{align*}

The results presented here provide statistical justification for our algorithm presented in the previous section where we select episodes based on their overlap score as given by ~\eqref{eq:os}. Suppose we have selected only $\alpha$ and want to choose either $\beta$ or $\gamma$ as our second episode. Based on ~\eqref{eq:os}, this choice depends on the sign of $(N-1)(f^{*}_{\beta} - f^{*}_{\gamma}) -  (O^{*}_{\alpha \beta} -O^{*}_{\alpha \gamma})$, which is a figure of merit motivated by considerations of coding efficiency. This is essentially same as the difference of $Overlap\text{-}score_1$ between $\beta$ and $\gamma$ which is a figure of merit that determines which pair of episodes maximize the data likelihood. 

\section{Application to Text Classification}
\label{sec:application}

In this section we present a novel application of our method of finding a `good' subset of frequent episodes to characterize data. The application is in the domain of text classification. Most text classification techniques use a bag-of-words approach where each document (or data sample) is represented as a collection of words that belong to a dictionary. The dictionary is usually considered as the set of all unique words present in the training corpus after preliminary preprocessing. This makes the size of the dictionary large leading to high dimensionality of the feature vector representation of each document. Other vector space representation of documents (e.g.,word-averaging in  \cite{socher2013recursive}) also, depends largely on the dictionary of words used.  %Larger size of dictionary leads to larger computational burden. 

  One can think of a text document as a sequence of events with event types being the words. Then using all training data in an unsupervised fashion, we can use our method to find the `best' subset of serial episodes that represent the data well. These episodes are likely to contain all specific words that are important for this document collection. 
% and some of the episodes may represent specially useful phrases. 
Thus, a dictionary built using only the words (event-types) found in  the subset of discovered episodes is likely to be useful. This is what we explore here. 

Let the dictionary obtained by using all unique words (after usual preprocessing) from the training data collection be termed \textbf{Dictionary-I}. 
We run our algorithm for discovering the `best' subset of serial episodes (that achieve best data compression) on the entire training corpus. We form a new dictionary as the set of all the unique words (event-types) that are present in the non-singleton episodes (i.e., episodes of size 2 or more) discovered by our algorithm. We call this smaller sized dictionary as \textbf{Dictionary-II}. 
%Apart from this, we consider another dictionary (\textbf{Dictionary-III}), where the non-singleton episodes discovered are added as extra `super-words' along with the words of \textbf{Dictionary-II}. 
In each case we would represent documents as vectors over one of these dictionaries and investigate standard classifiers such as Naive-Bayes and SVM. Using simulations on some standard benchmark datasets we show that we get large dimensionality reduction without any loss of accuracy by the classifier. 
%  In this section, we investigate the utility of local frequent patterns for text classification application. The CSC-2 algorithm is intended to obtain characterizing patterns from a single large sequence. 

%We recall that our method of mining for the subset of characterizing episodes does not need a frequency threshold or any other user-supplied parameter. Thus this is a parameterless (and unsupervised) method of dimensionality reduction for text classification problems. 

Typically, in training data for text classification, we have many documents but each document is short. Mining for episodes that can achieve significant compression individually for each document does not give any interesting episodes mainly because each sequence is short. We string together all training data (of all classes) to make one long document and we mine for a set of frequent serial episodes that achieve best compression (using the algorithm discussed in this paper). We employ special symbols to denote end of each training document and modify our mining algorithm so that occurrence of no episode would span two different documents. 

\subsection{Experimental Results}
%We show our experimental results on three different text datasets, where each document is represented in a bag-of-words model with the different dictionaries mentioned above. 
\subsubsection{Datasets}
We compared the classification accuracies on three standard benchmarks, {\em 20-Newsgroup}, {\em Reuters-21578} and {\em WebKB}, downloaded from a publicly available repository of datasets for single-label text 
categorization.\footnote{\label{footnote:1}\url{http://ana.cachopo.org/datasets-for-single-label-text-categorization}} We used the preprocessed stemmed version of these datasets. For {\em Reuters-21578}, we use the 8 class stemmed version of the dataset; {\em WebKB-} is a 4-class dataset while {\em 20-Newsgroup-} is a 20-class dataset. For these, the \textbf{Dictionary-I} is the set of all unique words present in the stemmed training data.
%In this website there is also a description of the datasets, steps taken to make each dataset single-labelled, their standard train/test splits, the pre-processing techniques applied to each dataset. 
Apart from these, we also used the movie-review dataset prepared by Pang and Lee (2004). We used the \textit{polarity dataset v2.0}.\footnote{\label{footnote:2}\url{http://www.cs.cornell.edu/people/pabo/movie-review-data/}} This sentiment analysis dataset consist of 2000 movie reviews. As preprocessing steps, we converted all letters to lower case and removed all words less than 3 characters long. No stop words except `and', `the' were removed. \textbf{Dictionary-I} was created from this preprocessed training data. %We present 10-fold cross-validation accuracy on the original folds introduced in \cite{pang_lee}.

%{\em 1.Reuters-21578}: We use the 8 class stemmed version of the dataset. 
%\textbf{Dictionary-I} was the set of all unique words present in the stemmed training data. 
%Apart from the original train-test split available in the given website, we generated three more random train-test splits keeping the train-test distribution of each class same as in the original split. 
%We present the mean classification accuracy on these 4 different splits.

%{\em 2.WebKB-} We use the stemmed version of this 4 class dataset. Like the previous dataset, similar procedures were followed to generate \textbf{Dictionary-I}.
%and three more random splits of the dataset. 

%{\em 3.20-Newsgroup-} This is a 20 class dataset on news articles. Similar steps like above were followed. 

%{\em 4.Movie Review-}This sentiment analysis dataset consist of 1000 positive and 1000 negative movie reviews. As preprocessing steps, we converted all letters to lower case and removed all words less than 3 characters long. No stop words except `and', `the' were removed. \textbf{Dictionary-I} was created from this preprocessed training data. We present 10-fold cross-validation accuracy on the original folds introduced in \cite{pang_lee}.

\subsubsection{Feature Vectors}
We compared the text classification accuracies using two different  models
\begin{itemize}
\item \textbf{Bag-of-words(BoW)}- Each data sample is converted into a feature vector of the dimension of the size of the corresponding dictionary used. Each feature represents the frequency of that word in that data sample except for the Movie Review dataset, where, as in \cite{pang_lee}, we used binary features denoting presence or absence of the word in the corresponding document. Further, tf-idf along with cosine normalization were done on these feature vectors as explained in the next subsection. 
\item \textbf{Average Embedding (VecAvg)} \cite{socher2013recursive}- Word2vec is used to produce the word embeddings and each text is then represented as the average of all the embeddings of the words present in that text. In case of \textbf{Dictionary-II}, averaging of word embeddings were done only for words which were part of the dictionary and the rest were ignored. In case of Movie Reviews and 20 Newsgroup, the pretrained model of GoogleNews vectors \footnote{\label{footnote:6} \url{https://github.com/mmihaltz/word2vec-GoogleNews-vectors}}
 were used, whereas in case of the other two datasets (since these were stemmed), the model was trained with \texttt{gensim} library with parameters \texttt{vector size=200}  and \texttt{window=5}.
\end{itemize}

\subsubsection{Tf-Idf}
Term  frequency-Inverse  document  frequency  (tf-idf) is a numerical statistic which is good at quantifying the importance of a word  to  a  document  in  a  collection. Let $wf(w,d)$ denote the frequency (that is the number of occurrences) of a word $w$ in a document $d$. Instead of using this raw frequency as the feature value, we use a modified word frequency defined by 
%This feature vector is then modified as described below. For a word $w$ in document $d$,
\begin{equation*}
Modified\text{-}wf(w,d) = wf(w,d)*idf(w)
\end{equation*}
 where the inverse-document frequency, ($idf(w)$), is given as 
\begin{equation*}
idf(w)=\log \frac{1+n_d}{1+df(w)} +1
\end{equation*}
Here, $n_d$ is the total number of documents and $df(w)$ is the number of documents that contain the word $w$. We use this modified frequency of each word ($Modified\text{-}wf(w,d)$) as the feature value. 
%(It may be noted that this is obtained using a variant of tf-idf). 
The feature vectors were further cosine normalized.

\subsubsection{Results}
We compare the classification accuracies obtained using our proposed \textbf{Dictionary-II}  with those obtained with \textbf{Dictionary-I}. For BoW and VecAvg representation, we present results using Linear SVM. For BoW, Naive Bayes(NB) results are also presented for comparison with accuracies reported in literature. For the Movie Review dataset, we present the mean value corresponding to the ten fold cross validation on the original folds introduced in \cite{pang_lee}. 
%For the Movie Review and 20-Newsgroup datasets, just to speed up the episode discovery process, we mined for only the top 8000 episodes  and constructed \textbf{Dictionary-II} from those episodes. 
%The inter-event gap ($T_g$) in CSC-2 was taken as 3 for all the cases. 
\begin{table}[h]
	\centering
	
	\begin{tabular}{ |c|c|c|c|} 
		\hline
		\textbf{Dataset} & \specialcell[c]{\bf Number of \\ \bf discovered \\ \bf episodes} & \specialcell[c]{\bf Size of \\ \bf Dict-I}& \specialcell[c]{\bf Size of \\ \bf Dict-II}  \\ \hline
		\textit{Reuters-21578} &  2261 & \quad 14575  & \quad 1560   \\ \hline
	\textit{WebKB} &  2423 & \quad 7287   & \quad 1884   \\ \hline
	\textit{20-Newsgroup} &  4703 & \quad 54580   & \quad 7361   \\ \hline
	\textit{Movie Review} &  2490 & \quad 37714  & \quad 3007   \\ \hline		
	\end{tabular}
\caption{Dictionary sizes for different datasets}
\label{table:2}
\end{table}

\begin{table}
\centering

\begin{tabular}{|c|c|c|c|c|}
\hline
\multirow{3}{*}{Dataset} & \multicolumn{4}{c|}{Scores ( \%)}                                                \\ \cline{2-5} 
                         & \multicolumn{2}{c|}{\textbf{Accuracy}} & \multicolumn{2}{c|}{\textbf{F-measure}} \\ \cline{2-5} 
                         & Dict-I       & Dict-II     & Dict-I       & Dict-II      \\ \hline
\textit{Reuters-21578}   & 95.43              & \textbf{95.52}    &   81.67     &    \textbf{81.97}         \\ \hline
\textit{WebKB}           & \textbf{89.11}     & 88.90             & \textbf{87.84}     & 87.63              \\ \hline
\textit{20-Newsgroup}    & 69.84              & \textbf{70.14}    & 68.23              & \textbf{68.64}     \\ \hline
\textit{Movie Review}    & \textbf{81.3}      & 80.35             &       \textbf{81.28}             &              80.34      \\ \hline
\end{tabular}

\caption{Linear SVM accuracy and F-measure(macro) for VecAvg representation}
\label{table:word2vec}
\end{table}

\begin{table}
\centering
\captionsetup{justification=centering}

\begin{tabular}{|c|c|c|c|c|c|}
\hline
\multirow{3}{*}{Dataset}       & \multirow{3}{*}{Classifier} & \multicolumn{4}{c|}{Scores (\%)}                                                             \\ \cline{3-6} 
                               &                             & \multicolumn{2}{c|}{Accuracy}                    & \multicolumn{2}{c|}{F-measure}            \\ \cline{3-6} 
                               &                             & Dict-I         & Dict-II               & Dict-I         & Dict-II      \\ \hline
\multirow{2}{*}{Reuters-21578} & NB                          & 96.07          & \textbf{96.30}          & 95.99          & \textbf{96.32}    \\ \cline{2-6} 
                               & SVM                         & 97.03          & \textbf{97.17}           & 97.01 &  \textbf{97.13}            \\ \hline
\multirow{2}{*}{WebKB}         & NB                          & 83.52          & \textbf{83.60}           & 82.26         & \textbf{83.63}    \\ \cline{2-6} 
                               & SVM                         & \textbf{91.04} & 90.62                  & 90.54 & \textbf{90.61}           \\ \hline
\multirow{2}{*}{20-Newsgroup}  & NB                          & \textbf{81.03} & 79.41                   & \textbf{79.89} & 79.33            \\ \cline{2-6} 
                               & SVM                         & \textbf{82.73} & 81.99                    & \textbf{82.03} & 81.90             \\ \hline
\multirow{2}{*}{Movie Review}  & NB                          & 82.50          & \textbf{82.85}           & \textbf{82.48}          & 82.34   \\ \cline{2-6} 
                               & SVM                         & \textbf{86.75}          & 84.50           & \textbf{87.64} & 84.74            \\ \hline
\end{tabular}

\caption{Naive-Bayes, Linear SVM Classification accuracy and F-measure(macro)for BoW representation}
\label{table:3}
\end{table}

Table \ref{table:2} shows sizes of the two dictionaries for  different datasets.
%original split of the two datasets whereas Table-\ref{table:3} reports the mean values of sizes corresponding to the four splits for the \textit{Reuters} and \textit{WebKB} datasets and ten folds for the \textit{Movie Review} dataset.
The number of episodes reported in Table~\ref{table:2} is  the number of non-singleton episodes. As can be seen from the table, the size of \textbf{Dictionary-II} is almost a fourth of that of \textbf{Dictionary-I} in case of \textit{WebKB}; for the other datasets it is about one eighth to one tenth. Thus, this method results in a very significant reduction in dictionary size (and hence in feature vector dimension).
%Below we present classification accuracies when documents are represented using the different dictionaries.
%\begin{table}[h]
%	\centering
%	\captionsetup{justification=centering}
%	\begin{tabular}{ |c|c|c|c| } 
%		\hline
%		\textbf{Dataset} & \textbf{\textbf{Dictionary-I}} & \textbf{ \textbf{Dictionary-II}} & \textbf{ \textbf{Dictionary-III}} \\ \hline
%		\textit{Reuters-21578} & \quad 96.07,97.03  & \quad \textbf{96.25,97.21}  & \quad 95.98,97.12  \\ \hline
%		\textit{WebKB} & \quad 83.52,\textbf{91.04} & \quad 83.09,90.54  & \quad \textbf{85.31},90.76  \\ \hline
%	\end{tabular}
%	\caption{Naive-Bayes, Linear SVM Classification accuracy using different dictionaries for the original split }
%	\label{table:4}
%\end{table}

\begin{table}[h!]
	\centering
	\captionsetup{justification=centering}
	\begin{tabular}{ |c|c|c| } 
		\hline
		\textbf{Dataset} & \textbf{\textbf{Dictionary-I}} & \textbf{ \textbf{Dictionary-II}}  \\ \hline
		\textit{Reuters-21578}& 95.36($\pm$0.883) & 95.01($\pm0.895$)   \\ \hline
		\textit{WebKB} & 83.38($\pm$0.315) & 83.23($\pm$0.328)    \\ \hline
		\textit{20-Newsgroup} &  81.35($\pm0.412$) & 80.36($\pm0.532$)  \\ \hline	
	\end{tabular}
	\caption{Mean (and standard deviation) of classification accuracy with Naive-Bayes using different dictionaries (BoW representation)}
\label{table:4}
\end{table}

\begin{table}[h!]
	\centering
	\captionsetup{justification=centering}
	\begin{tabular}{ |c|c|c| } 
		\hline
		\textbf{Dataset} & \textbf{\textbf{Dictionary-I}} & \textbf{ \textbf{Dictionary-II}}  \\ \hline
		\textit{Reuters-21578}  &  97.30($\pm$0.289) & 97.31($\pm$0.214)   \\ \hline
		\textit{WebKB}  & 89.97($\pm$0.798) & 90.58($\pm$0.536)     \\ \hline
		\textit{20-Newsgroup} & 82.94($\pm0.139$)  & 82.67($\pm$0.473)    \\ \hline	
	\end{tabular}
	\caption{Mean (and standard deviation) of classification accuracy with Linear SVM using different dictionaries (BoW representation)}
	\label{table:5}
\end{table}

The classification accuracies obtained with different dictionaries are shown in Tables~\ref{table:word2vec}--\ref{table:3}. Table~\ref{table:word2vec} shows accuracies and F-measure with linear SVM classifier under Vec Avg representation while Table~\ref{table:3} shows these for Naive Bayes and linear SVM classifiers under BoW representation.  We did not try any nonlinear SVM because all other studies on these benchmark data sets reported only accuracies with linear SVM. As is easy to see, the accuracies  and F-measure scores (under both BoW as well as VecAvg representation) achieved by either classifier with different dictionaries are mostly very close. 
%The mean accuracies are within one standard deviation of each other in most cases. 
 Thus we can conclude that our frequent episodes based method allows us to get a very large reduction in dictionary size without any significant change in the classification accuracy. (We also note that 
the accuracy of our \textbf{Dictionary-I} in Table \ref{table:3} is consistent with the bag-of-words accuracy reported in \cite{textdata_1} and \cite{wordvectors_2011}). 
%Even for VecAvg representation using Word2Vec, we see the scores are comparable using the default and our dictionary (Dictionary-II). 

The above are with the train-test split as given in the original datasets. For BoW representation, we also generated 3 random splits for the datasets \textit{Reuters-21578, WebKB, 20-Newsgroup} having the same train-test distribution of each class as in the original split. The results (showing averages and standard deviations) are presented in Tables~\ref{table:4}--\ref{table:5}. Once again, the results clearly show that there are no significant differences between accuracies achieved with the two dictionaries.

\begin{table*}[t]
\centering
\captionsetup{justification=centering}
\resizebox{\textwidth}{!}{\begin{tabular}{|c|l|l|l|l|}
\hline
\multirow{2}{*}{}                                                  & \multicolumn{4}{c|}{Dataset}                                                                                                                                                                                                                                                                                                                                                                                                                                                                                                                                                                                                                                                                                                                                                                                                                                                                                                                                                            \\ \cline{2-5} 
                                                                   & \multicolumn{2}{c|}{\begin{tabular}[c]{@{}c@{}}\textbf{Movie Review}\\ labels=`positive sentiment',              `negative sentiment'\end{tabular}}                                                                                                                                                                                                                                           & \multicolumn{2}{c|}{\begin{tabular}[c]{@{}c@{}}\textbf{WebKb}\\ labels=`project',`student', `faculty',`course'\end{tabular}}                                                                                                                                                                                                                                                                                                                                                                                                                                                                          \\ \hline
\textit{\begin{tabular}[c]{@{}c@{}}Rejected \\ Words\end{tabular}} & \multicolumn{2}{l|}{\begin{tabular}[c]{@{}l@{}}stunts, theatre, cinematographer, moviestar,\\directorship, producers, storyteller, scripts, \\ spotlight,   audition, auditorium,  backstage, \\torrent, reviewer, performances, entertainment.\end{tabular}} & \multicolumn{2}{l|}{\begin{tabular}[c]{@{}l@{}}chemistry, cryptography,  probabilistic, lagrangian, \\ arithmetic,  scholarship,  bibtex, manuscript, newsletter,\\  computer, interdisciplinary, mathematician, biotechnology, \\accuracy, baseline, neurocomputing, gaussian.\end{tabular}}                                                                                                                                                          \\ \hline
\textit{\begin{tabular}[c]{@{}c@{}}Selected \\ Words\end{tabular}} & \multicolumn{2}{l|}{\begin{tabular}[c]{@{}l@{}} enjoyable, funny, hilarious, entertaining,\\ superb, boring,  sleepy,  disappointed,  twists,\\ clever, impressed, surprised, liked, interested, \\awful, pleasing, miserably,  dumber, interesting, \\impressive, intelligent, fantastic.\end{tabular}}                                      & \multicolumn{2}{l|}{\begin{tabular}[c]{@{}l@{}}syllabus, internet,  introductory, prerequisite, research,\\  bibliography, professor, student, quiz, exercise, credit,  \\ query, tutor,  project,  phd, fellowship, conference,\\ curriculum, scientist, magazine, instructor, theorem,\\ homework, examination, semester, journal, homepage. %\\technology, abstract,  algorithm, workshop, springer .
\end{tabular}} \\ \hline
\end{tabular}}
\caption{Sample words from the set of rejected and selected words for Dictionary-II}
\label{table:word_list}
\end{table*}

For the BoW representation for this document classification application, our method of learning a dictionary results in a significant decrease in feature vector dimension. But this method is quite different from generic dimensionality reduction techniques such as PCA. With PCA we may get dimensionality reduction by choosing certain linear combinations of earlier features. With the original feature vector dimension being in tens of thousands, the new features obtained as such linear combinations would not be semantically interpretable. However, our data mining method essentially decides on which words of \textbf{Dictionary-I} to be retained (and which are to be rejected). Thus this method is essentially a feature selection method rather than a dimensionality reduction method. Hence, the dimensionality reduction achieved here is semantically interpretable. 

 To get such a feel for what the data mining does, we 
present in Table-\ref{table:word_list}, some sample of words that are retained and rejected by our method in case of Movie Review and WebKB datasets. The words shown  are hand-picked but only from a set of 1000 randomly selected words. It is easy to see that this makes good semantic sense. For example, in Movie Review we reject many movie related words like `stunts', `theater', `performances' etc. which, while they may appear in the reviews,  may not carry any information regarding sentiment of the review.  On the other hand, we retain words like `hilarious', `boring', 
`surprised' etc. that can carry sentiment information. Similar comments apply to WebKb dataset (e.g., selected words like `prerequisite', `introductory', `project' can be commonly found on a project or course web-page and hence they may carry discriminative information). Thus, the data mining method (based on finding episodes for compressing data) seems to be effective in picking a dictionary that is relevant to the text corpus.

\section{Conclusions}
\label{sec:conclusions}

In this paper we considered the problem of discovering a small set of serial episodes to characterize sequential data. We extended the existing CSC-2 algorithm of ~\cite{ibrahim_episodes} to work with non-overlapped frequency. 
%This algorithm is motivated based on the MDL principle. 
%Using an intuitively appealing coding scheme to encode data using episodes, the algorithm finds a subset of episodes to maximize data compression achieved.

Our main contribution is a novel HMM-based generative model for pairs of episodes. 
The model generates very general output sequences where the two episodes are the most prominent frequent episodes (under non-overlapped frequency). 
The model is very intuitive. The symbols emitted from episode states constitute the `model-based' occurrences of episodes. The noise states can emit any symbol and hence symbols emitted from the noise states can be thought of as the distracting signal that may mask real episodes and contribute spurious frequent episodes. The transition structure is also intuitively motivated.  From any state, transitions into a noise state has probability $\eta$. The remaining probability is equally divided between all reachable episode states. For this model class we showed that the episode-pair model that has best likelihood for the data sequence, is determined both by the frequencies of the episodes as well as overlaps between their occurrences. The analytical expressions we derived for the data likelihoods provide statistical justification for our algorithm of selecting a subset of episodes. 

The CSC-2 algorithm is motivated based on the MDL principle. 
Using an intuitively appealing coding scheme to encode data using episodes, the algorithm finds a subset of episodes to maximize data compression achieved. It is essentially incrementally picking episodes based on the so called overlap score which depends both on the frequency of the episode as well as on the extent of overlap in its occurrences with those of already selected episodes. Our HMM-based model provides some statistical justification for this strategy used by the algorithm. 

A generative model for sequential data to capture interactions of two episodes as well as using it to justify an MDL based algorithm for frequent episodes are both novel contributions of this paper. As mentioned in Section~1, there have been many algorithms, motivated by the MDL philosophy, for succinctly characterizing data using a small set of frequent patterns. However, all such algorithms for sequential data are heuristic in nature. We believe that the HMM model we presented here is a good first step in developing a statistical theory for MDL-based algorithms that find a good subset of frequent episodes.

Another important contribution of this paper is a novel application of frequent episodes mining to text classification. We view the text document as a sequence of events with event types being the words. Then we find the subset of episodes that best characterizes the entire text corpus in terms of data compression. The words appearing in this subset of frequent episodes is likely to gives us the most informative words for the corpus and hence we use only these words to form the dictionary using which the documents are represented as vectors. Thus the method amounts to learning a context-sensitive dictionary using the idea of frequent pattern mining. Also, since our data mining method does not need any user-specified hyperparameters, same is true for this method of dimensionality reduction. To the best of our knowledge this is a first instance of application of frequent pattern methods to dictionary learning. 
As we showed through extensive simulations, the method results in many-fold decrease in the size of dictionary without compromising the classification accuracy. Also, as can be seen from the examples of retained and rejected words, the method seems to be quite effective in learning a good subset of words. 

The HMM model we presented is for pairs of episodes. While it is, in principle, extendable to any number of episodes, notationally it would be very complex. A good extension of the work presented here is in the direction of extending these analytical techniques to arbitrary number of episodes. Generative models can, in general, be used for assessing statistical significance of the frequency of an episode (e.g., \cite{laxman_hmm}). Since the model introduced here also accounts for interactions among episodes, it should be usable for questions such as whether or not the observed frequencies of two episodes  would make both of them significant given the extent of overlap between their occurrences. This is also a useful direction in which the work presented here can be extended. 

\ifCLASSOPTIONcaptionsoff
  \newpage
\fi



%\bibliographystyle{IEEEtran}
\newpage
\bibliographystyle{IEEEtran}
\bibliography{references}

\vfill

%\newpage



%\vfill


\end{document}